\documentclass[11pt]{article}

\usepackage[final]{acl}

\usepackage{times}
\usepackage{latexsym}

\usepackage[T1]{fontenc}

\usepackage[utf8]{inputenc}

\usepackage{microtype}

\usepackage{inconsolata}
\usepackage{longtable}
\usepackage{changepage}
\usepackage{float}
\usepackage{makecell}
\usepackage{array}
\usepackage{enumitem}
\usepackage{graphicx}
\usepackage{dsfont}
\usepackage{fontawesome}
\usepackage{titletoc}
\usepackage{colortbl}
\usepackage{multirow}
\usepackage[most]{tcolorbox}
\usepackage{amssymb}
\usepackage{booktabs} 
\usepackage[table]{xcolor}
\usepackage[normalem]{ulem}
\usepackage{array}
\usepackage{hhline}
\usepackage{soul}

\soulregister\citep7
\soulregister\citet7
\soulregister\ref7

\newcommand{\softscore}[1]{\color[HTML]{999999} \scriptsize (#1)}
\newcommand{\checkbox}[1]{%
  \ifnum#1=1
    \makebox[0pt][l]{\raisebox{0.15ex}{\hspace{0.1em}$\checkmark$}}%
  \fi
  $\square$%
}
\newcommand{\globalbest}[1]{\cellcolor[HTML]{D3E8EE}\textbf{#1}}
\newcommand{\groupbest}[1]{\uline{#1}}

\title{PUMA: A Polish Benchmark for Culturally Grounded Multimodal Understanding}

\author{
\textbf{Sławomir Dadas},
\textbf{Michał Perełkiewicz},
\textbf{Rafał Poświata},
\textbf{Małgorzata Grębowiec}, \\
\textbf{Bartłomiej Jaworski},
\textbf{Izabela Woźniakowska} \\
National Information Processing Institute \\
al. Niepodległości 188b, 00-608 Warsaw, Poland}

\begin{document}
\maketitle
\begin{abstract}
Large language models are increasingly moving beyond text processing, adding support for other modalities such as images and audio. While text understanding and generation have been extensively studied, multimodal data processing capabilities, particularly in the context of cultures and languages other than English, have not yet been evaluated comprehensively. In this paper, we propose PUMA (Polish Unified Multimodal Assessment), a novel benchmark of 900 hand-crafted tasks designed to probe the limits of multimodal models in the Polish cultural and linguistic context. The dataset evaluates both cultural understanding and practical skill in processing text, images, audio, and visually rich documents. Our extensive evaluation of frontier commercial models, open-weights models, and specialized smaller systems highlights a significant performance gap. While top commercial models achieve high scores in visual question answering, most models struggle with complex audio or document understanding. We open-source our evaluation framework to advance localized multimodal AI research.
\end{abstract}

\section{Introduction}

In recent years, the range of capabilities of large language models has expanded significantly. Systems that were initially designed primarily for text processing and generation increasingly support other modalities as well, such as images, documents, audio, and video. At the same time, their ability to operate across many natural languages continues to improve, as does their capacity to understand and generate programming languages and other formal languages. However, mastery of grammar, vocabulary, or discourse structure alone is not sufficient for full language understanding. The meaning of many utterances depends on a broader cultural context: references to history, traditions, literature, art, social life, or contemporary popular culture \citep{hershcovich2022challenges}. A lack of familiarity with these nuances may lead to superficial or incorrect interpretations, even when a model correctly identifies the literal meaning of the text.

For this reason, increasing attention is being paid to evaluating models in terms of their understanding of cultural context. In recent years, benchmarks have emerged that examine the extent to which models can recognize and interpret cultural references \citep{pawar2025survey}. Initially, these benchmarks focused mainly on text-based tasks. However, with the development of multimodal models, it is becoming increasingly important to assess whether such systems can combine linguistic information with images, documents, or other media.

In this paper, we present \textbf{PUMA} (\textbf{P}olish \textbf{U}nified \textbf{M}ultimodal \textbf{A}ssessment) - a benchmark consisting of \textbf{900 hand-crafted tasks} designed to evaluate the linguistic and cultural competencies of large models in the context of Poland and the Polish language. The tasks span three modalities: images, audio, and documents. Each modality contains three categories targeting distinct areas of knowledge or practical competencies. The benchmark includes both culturally grounded question-answering tasks and tasks assessing practical capabilities in multimodal data processing. To the best of our knowledge, PUMA is the first Polish-language benchmark to jointly evaluate image, audio, and document understanding within a single framework, combining culturally grounded and practical tasks such as ASR, OCR, and structured information extraction. Our contributions are as follows:
\begin{itemize}[wide,labelwidth=0pt,labelindent=0pt,itemsep=0pt,topsep=5pt,parsep=0pt]
\item We introduce a new Polish-language benchmark for evaluating large multimodal models.\footnote{\url{https://huggingface.co/spaces/OPI-PIB/puma}} The dataset was manually crafted by a team of annotators and encompasses tasks related to processing text, images, audio, and documents.
\item We evaluate over a dozen open-weight and commercial multimodal models and over 50 additional visual language models (VLMs) on the visual tasks.
\item We compare large multimodal models with smaller specialized models on automatic speech recognition (ASR) and optical character recognition (OCR).
\item As part of the work on the benchmark, we developed software for building multimodal question datasets and running evaluations. We are releasing this tool as open-source.\footnote{\url{https://github.com/OPI-AILab/puma-app}}
\end{itemize}

\begin{figure*}
  \centering
  \includegraphics[scale=0.85]{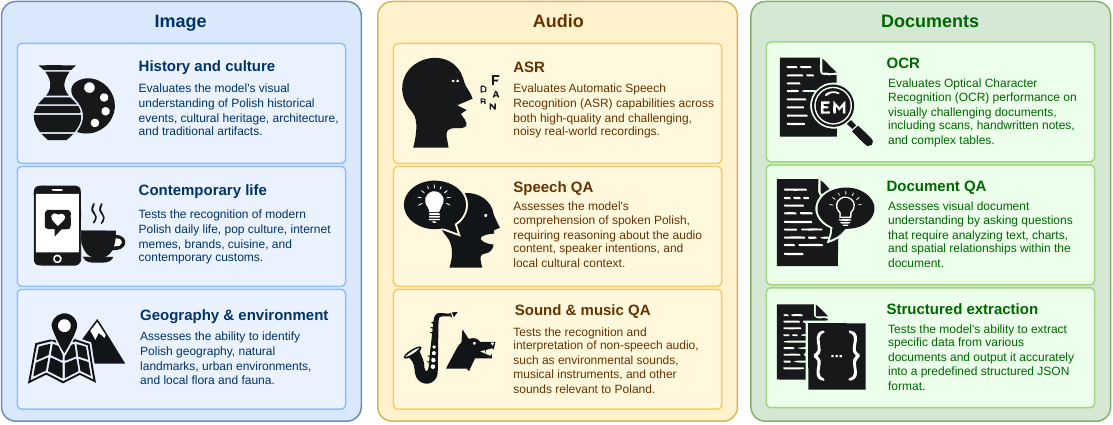}
  \caption{A diagram showing the breakdown of categories in our benchmark. The nine categories, representing tasks types or subject areas, span three modalities: images, audio, and documents.}
  \label{fig:categories}
\end{figure*}

\section{Related Work}
Current evaluation methods for large language models generally span four main categories, starting with deterministic verification of closed-ended tasks such as multiple-choice questions \citep{NEURIPS2024_ad236edc}, math problems \citep{hendrycks2021measuring}, and classic NLP tasks like  binary or multi-label classification \citep{KOCON2023101861}. To assess open-ended generation, researchers frequently employ either the cost-intensive but highly aligned human evaluation \citep{10.5555/3692070.3692401} or the increasingly popular LLM-as-a-judge approach \citep{10.5555/3666122.3668142}, which scales well despite potential evaluator biases \citep{ye2025justice, shi-etal-2025-judging}. Additionally, programmatic evaluations apply predefined rules to accurately validate specific capabilities, such as code execution \citep{zhuo2025bigcodebench} or strict instruction following \citep{Zhou2023InstructionFollowingEF, suzgun-etal-2023-challenging}. However, as models achieve global adoption, these foundational evaluation paradigms have necessarily expanded to assess linguistic and cultural competencies extending beyond English and Western contexts.

Consequently, multiple initiatives have emerged to evaluate these cross-cultural abilities. BLEnD \citep{myung2024blend} evaluates everyday knowledge across diverse cultures and languages, CulturalBench \citep{chiu-etal-2025-culturalbench} assesses cultural knowledge spanning 45 global regions, and CDEval \citep{wang-etal-2024-cdeval} targets specific cultural dimensions. Additionally, SeaEval \cite{wang-etal-2024-seaeval} and MAPS \cite{liu-etal-2024-multilingual} focus on verifying cultural reasoning. Other notable multilingual benchmarks include CultureAtlas \citep{fung2024}, CultureBank \citep{shi2024culturebank}, CULTURE-GEN \citep{li2024culturegen}, and MultiNativQA \citep{hasan-etal-2025-nativqa}. This shift has also driven the creation of benchmarks tailored to specific cultures, such as FoundaBench \citep{li2024foundabench} for Chinese fundamental capabilities, Palm \citep{alwajih-etal-2025-palm} and AraDiCE \citep{mousi-etal-2025-aradice} for Arabic dialects and culture, and IndoCulture \citep{koto-etal-2024-indoculture} for regional Indonesian cultures. Similar region-specific datasets include BertaQA \citep{etxaniz2024bertaqa} for Basque, LaoBench \citep{Gao2025LaoBenchAL} for Lao, RoCulturaBench \citep{masala-etal-2024-vorbesti} for Romanian, KoBBQ \citep{jin-etal-2024-kobbq} for identifying Korean social biases. In Polish, the PLCC \citep{dadas2025evaluating} benchmark was developed to measure linguistic and cultural competencies using a rule-based approach. For evaluating QA categories, PUMA uses a rule-based approach inspired by this benchmark, but extended to support multimodality.

Extrapolating beyond text, the rapid development of multimodal models has introduced tasks requiring the interpretation of culturally grounded multimedia. Multimodal visual benchmarks such as CulturalVQA \citep{nayak-etal-2024-benchmarking}, SEA-VQA \citep{urailertprasert-etal-2024-sea}, ALM-bench \citep{Vayani_2025_CVPR}, and CVQA \citep{mogrovejo2024cvqa} test how well visual language models (VLMs) understand culturally diverse content, while BLEnD-Vis \citep{tan-etal-2026-blend} evaluates robustness across linguistic rephrasings and visual modalities. Further expanding this scope, CUBE \citep{kannen2024beyond} and IndicTTI \citep{mittal2024indicTTI} verify competence and biases in text-to-image generation, CROPE \citep{nikandrou-etal-2025-crope} evaluates in-context adaptation to cultural concepts, and platforms like MIXCUBE \citep{kim-etal-2025-tom} and M5 \citep{schneider-sitaram-2024-m5} examine awareness within cultural mixture contexts. Region-specific multimodal paradigms have parallelly flourished, featuring CVLUE \citep{10.1609/aaai.v39i8.32884} for Chinese VLMs, K-Viscuit \citep{park-etal-2025-evaluating} for Korean culture, HaVQA \citep{parida-etal-2023-havqa} for Hausa. Broader audio and multimodal benchmarks also exist, such as IndicVoices \citep{javed-etal-2024-indicvoices} for Indian languages and SEACrowd \citep{lovenia-etal-2024-seacrowd} for Southeast Asian languages. 

For Polish, PCVB\footnote{\url{https://hf.co/spaces/speakleash/Polish_Cultural_Vision_Benchmark}} assesses visual knowledge of Polish cultural heritage, history, geography, and traditions. Limited documentation prevents a detailed comparison, but available descriptions suggest that PCVB mainly requires recognizing objects, landmarks, foods, or people and identifying their country of origin, whereas PUMA uses more varied task types. The second benchmark for evaluating vision models in Polish is the Polish version of MMBench \citep{statkiewicz2026annotation}. It is a collection of questions translated from English to Polish and therefore focuses on general knowledge rather than knowledge grounded in Polish culture.

Compared to benchmarks described above, PUMA covers three modalities and features a wide range of tasks, from culturally grounded visual and speech QA, through language-specific OCR and ASR tasks, to real-world use cases requiring document understanding and structured information extraction.

\section{PUMA}
PUMA comprises 900 hand-crafted tasks assessing Polish cultural and linguistic competencies across images, audio, and documents. Each modality contains three 100-task categories, for nine categories in total. Six use question answering to evaluate knowledge and reasoning; the remaining three test practical skills: automatic speech recognition (ASR), optical character recognition (OCR), and structured extraction from documents. Evaluation is deterministic and requires no external judge \citep{Zhou2023InstructionFollowingEF}, making it fast and helping keep costs manageable when using commercial models.

The questions were created using a custom application developed specifically for this benchmark. It serves as a question database, enabling users to search, add and edit questions, integrate with models, verify answer correctness, run evaluations, and review the evaluation logs. More detailed information about this tool is provided in Appendix \ref{app:application}. Once the question creation process was completed, it was followed by a review and revision phase. To this end, we ran an evaluation using selected frontier multimodal models. Annotators were tasked with reviewing the logs from these runs and raising any issues regarding the question content or verification criteria, which were subsequently discussed. As a result of this phase, 28 questions were removed and replaced with new ones, while 57 underwent major modifications, such as rewriting the question prompt or changing the verification rules. Minor editorial fixes and typo corrections were made continuously during the review process. The annotation and verification processes are described in detail in Appendix~\ref{app:annotation_process}.

\subsection{Categories}
This section describes the scope of the categories included in our benchmark. The category structure is also illustrated in Figure \ref{fig:categories}. Here, we provide basic information about each category. A detailed descriptions, including instructions for annotators and question statistics, are provided in Appendices \ref{app:image_categories}, \ref{app:audio_categories}, and \ref{app:document_categories}.

\subsubsection{Image categories}

\noindent \textbf{History and culture} - This category tests the models' knowledge regarding the history, tradition, cultural heritage, and customs of Poland. It also includes art, primarily painting, sculpture, and architecture. Some of the questions concern people or objects of significant historical importance.

\noindent \textbf{Contemporary life} - It covers contemporary life and pop culture. Some of the questions involve modern media such as film, television, and the internet. We have also included questions related to sports, contemporary politics, and show business. Cultural and social issues are also addressed, provided they relate to contemporary topics.

\noindent \textbf{Geography and environment} - This category verifies the models' knowledge of Poland's geography, including its fauna, flora, and well-known natural landmarks. Questions about man-made structures may also be found in this category. These are primarily related to infrastructure, cities, as well as administrative and socioeconomic issues.

\subsubsection{Audio categories}

\noindent \textbf{Automatic Speech Recognition (ASR)} – This is a specific category designed to test models’ ability to transcribe speech, a task of significant practical importance. The recordings used in this category range from about a minute to several minutes in length. For the most part, these are challenging samples containing interference, background noise, complex language, or dialects.

\noindent \textbf{Speech QA} - It evaluates advanced speech understanding and knowledge of the Polish cultural context based on audio recordings. This category includes questions about short audio samples lasting from about a minute to several minutes. The scope of tested competencies covers speech comprehension, information extraction, recognition of the speakers' intentions and emotions, as well as the ability to link the content of the recording with broader knowledge about Poland.

\noindent \textbf{Sound and music QA} - This category includes questions about audio recordings where the main content is not speech, but other sounds such as ambient noises, musical instruments, animals, jingles, and the sounds of tools and machinery. All recordings are related to Poland, and the models' task is to recognize the sounds and use broader knowledge to interpret their content.

\subsubsection{Document categories}

\noindent \textbf{Optical character recognition (OCR)} - This category evaluates the models' ability to convert documents from image formats into plain text. The tasks mostly involve challenging cases, such as handwritten text or low-quality scans. Additionally, some examples contain tables. The metric we use verifies the correct extraction of tables while preserving their structure.

\noindent \textbf{Document QA} - This category tests the models' ability to understand and interpret documents in Polish by answering questions about their content. Most of the examples in this category consist of visually rich documents that require the analysis of text, tables, charts, or infographics, among others.

\noindent \textbf{Structured extraction} - This category evaluates the models' ability to extract relevant information from documents and save it in a structured JSON format. Each example in this category includes a document in the form of an image and defines a JSON schema that specifies the expected format of the model's response.

\subsection{Grading}

Questions are evaluated using verification rules. For each question, one or multiple such rules can be defined. A single rule verifies the alignment of the model's response with user-defined correctness criteria. To standardize the evaluation framework across tasks and rule types, we adopted a binary primary metric, meaning the model receives either zero or one point for its response. We refer to this as the \textbf{strict score}. Furthermore, for each rule, we defined a \textbf{soft score}, which yields fractional values if the model's response at least partially satisfies the predefined criteria. Since soft score distributions vary across different rule types, these values should be interpreted only as an auxiliary metric. If a question is associated with multiple verification rules, the response is deemed correct only if all rules are satisfied. The final score on the benchmark is calculated as the mean of the strict scores, representing the percentage of correct responses.

Our evaluation framework supports seven types of verification rules. Four of them are applicable to categories related to reasoning and question answering. Additionally, we introduced three specialized metrics to evaluate categories corresponding to other task types.

\subsubsection{Verification rules for question answering}
Correctness verification in the question answering categories relies primarily on matching characters, words, and phrases within the model's response. In addition to the prompt, the annotator specifies the target information that should or should not appear in the model's output, as well as, for certain questions, the expected sequence of these elements. Given that Polish is a morphologically rich language, this type of verification requires text normalization. By default, prior to applying the rules, all characters except letters and digits are removed from the text, all letters are converted to lowercase, and each word is reduced to its base form using a lemmatizer. If normalization should not be applied to a specific rule, the annotator can disable it. The following rules are available:

\noindent \textbf{Include} – Verifies whether a specified list of words or phrases is present in the model's response. In its simplest form, the rule accepts a comma-separated list of strings, requiring each phrase to appear in the response. It also allows for formulating expressions with logical disjunctions. In such cases, the condition is met if at least one of the phrases defined in the expression is successfully matched. Consequently, this rule supports arbitrary string-based expressions in conjunctive normal form (CNF). All clauses must be satisfied for the strict score to equal 1. The soft score for this rule is calculated as the percentage of correctly matched expressions. Optionally, the rule accepts additional parameters, \emph{include\_min} and \emph{include\_max}, which modify its default behavior. If provided, the rule is considered satisfied only if the number of matched expressions is at least \emph{include\_min} and no greater than \emph{include\_max}.

\noindent \textbf{Exclude} – Ensures that specific words or phrases defined in the condition are absent from the model's response. The response is classified as correct strictly if none of the specified phrases appear in the text. The soft score for this rule is computed as the percentage of phrases successfully excluded.

\noindent \textbf{Regex} – Verifies the presence of a specified regular expression within the model's response. By default, the rule is satisfied if at least one substring matches the regular expression. If boundary parameters are specified, the response must contain between \emph{regex\_min} and \emph{regex\_max} occurrences of the pattern. Additionally, it is possible to impose constraints on the length of the matched substrings and enforce that the matched strings constitute valid Polish words.

\noindent \textbf{Order} – Evaluates whether the specified words or phrases appear in the expected order within the response. This rule is particularly applicable to prompts requiring the model to generate a list, sort elements, or map names to specific objects. Furthermore, it is applicable for prompts containing multiple sub-queries or a sequence of true/false statements. The response is deemed correct strictly if all required phrases are present and appear in the correct sequential order. The soft score for this rule is computed using the normalized Kendall Tau distance: $K_n = \frac{\sum_{i=1}^{N-1} \sum_{j=i+1}^{N} \hat{o}(o_i, o_j)}{\frac{1}{2} N (N-1)}$, 
where the reference order of phrases $[o_1, o_2, ..., o_N]$ is given, and the function $\hat{o}(o_i, o_j)$ equals 1 if and only if both phrases $o_i, o_j$ occur in the model's response and $o_i$ occurs before $o_j$.

\begin{table*}[!t]
  \centering
  \small
  \setlength{\tabcolsep}{0.2pt}
  \renewcommand{\arraystretch}{1.04}

  \begin{tabular}{l|>{\centering\arraybackslash}m{1.1cm}|>{\centering\arraybackslash}m{1.1cm}|>{\centering\arraybackslash}m{1.1cm}|>{\centering\arraybackslash}m{1.1cm}|>{\centering\arraybackslash}m{1.1cm}|>{\centering\arraybackslash}m{1.1cm}|>{\centering\arraybackslash}m{1.1cm}|>{\centering\arraybackslash}m{1.1cm}|>{\centering\arraybackslash}m{1.1cm}!{\vrule width 1.06pt}>{\centering\arraybackslash}m{1.4cm}}
    \toprule
    & \multicolumn{3}{c|}{\textbf{Images}} & \multicolumn{3}{c|}{\textbf{Audio}} & \multicolumn{3}{c!{\vrule width 1.06pt}}{\textbf{Documents}} & \\
    \Xcline{2-10}{0.4pt}
    \multirow{-2}{*}{\textbf{Model}} & {\scriptsize History \& culture} & {\scriptsize Contemp. life} & {\scriptsize Geography \& env.} & {\scriptsize ASR} & {\scriptsize Speech QA} & {\scriptsize Sound \& music QA} & {\scriptsize OCR} & {\scriptsize Document QA} & {\scriptsize Structured extraction} & \multirow{-2}{*}{\makecell[c]{\textbf{Mean} \\ \textbf{score}}} \\[-1pt]
    \midrule

\multicolumn{11}{l}{\textbf{Multimodal models}}\\[-2pt]
\midrule

Phi-4-Multimodal-Instruct & 2 \softscore{6.2} & 2 \softscore{3.4} & 5 \softscore{11.8} & 0 \softscore{2.2} & 0 \softscore{2.2} & 2 \softscore{4.0} & 0 \softscore{0.6} & 6 \softscore{11.8} & 0 \softscore{14.8} & 1.9 \softscore{6.3} \\
MiMo-V2-Omni & 21 \softscore{30.6} & 27 \softscore{33.5} & 28 \softscore{42.9} & 3 \softscore{48.6} & 14 \softscore{28.5} & 23 \softscore{28.0} & 57 \softscore{82.2} & 53 \softscore{68.1} & 16 \softscore{39.5} & 26.9 \softscore{44.6} \\
MiMo-V2.5 & 22 \softscore{30.6} & 23 \softscore{32.5} & 31 \softscore{42.0} & 1 \softscore{49.1} & 15 \softscore{29.4} & 22 \softscore{28.4} & 57 \softscore{82.8} & 62 \softscore{72.1} & 39 \softscore{88.7} & 30.2 \softscore{50.6} \\
Gemini-2.5-Flash & 50 \softscore{58.2} & 40 \softscore{51.0} & 53 \softscore{66.2} & 69 \softscore{89.7} & 41 \softscore{52.2} & 23 \softscore{29.8} & 73 \softscore{91.0} & 52 \softscore{64.4} & 53 \softscore{91.9} & 50.4 \softscore{66.0} \\
Gemini-3.1-Flash-Lite & 58 \softscore{63.2} & 49 \softscore{58.2} & 62 \softscore{76.2} & 70 \softscore{87.1} & 35 \softscore{47.2} & 31 \softscore{38.6} & 76 \softscore{91.1} & 44 \softscore{60.0} & 45 \softscore{91.0} & 52.2 \softscore{68.0} \\
Gemini-2.5-Pro & 70 \softscore{76.0} & 72 \softscore{80.2} & 72 \softscore{80.8} & \globalbest{\groupbest{81} \softscore{91.5}} & 65 \softscore{76.1} & 45 \softscore{53.0} & 80 \softscore{92.3} & 72 \softscore{83.6} & 68 \softscore{96.4} & 69.4 \softscore{81.1} \\
Gemini-3.5-Flash & 82 \softscore{85.8} & 79 \softscore{85.3} & \globalbest{\groupbest{86} \softscore{92.6}} & 77 \softscore{89.2} & 81 \softscore{86.3} & 47 \softscore{51.5} & \groupbest{86} \softscore{94.3} & \globalbest{\groupbest{88} \softscore{94.7}} & 66 \softscore{94.9} & 76.9 \softscore{86.1} \\
Gemini-3.1-Pro & \globalbest{\groupbest{82} \softscore{86.9}} & \globalbest{\groupbest{83} \softscore{87.3}} & 85 \softscore{91.1} & 77 \softscore{91.0} & \globalbest{\groupbest{91} \softscore{93.5}} & \globalbest{\groupbest{56} \softscore{59.7}} & 81 \softscore{91.4} & 83 \softscore{89.7} & \globalbest{\groupbest{79} \softscore{97.3}} & \globalbest{\groupbest{79.7} \softscore{87.5}} \\

\midrule
\multicolumn{11}{l}{\small{\textbf{Model pairs (vision + audio)}}}\\[-2pt]
\midrule

Mistral-3.2-24B + Voxtral-24B & 15 \softscore{21.9} & 17 \softscore{24.9} & 7 \softscore{19.9} & 36 \softscore{80.3} & 23 \softscore{35.4} & 4 \softscore{12.5} & 1 \softscore{1.1} & 18 \softscore{36.5} & 18 \softscore{78.8} & 15.4 \softscore{34.6} \\
GPT-4o + GPT-4o-Audio & 41 \softscore{50.7} & 39 \softscore{47.2} & 48 \softscore{58.9} & 64 \softscore{82.6} & \groupbest{50} \softscore{59.3} & \groupbest{34} \softscore{40.0} & 17 \softscore{31.8} & 23 \softscore{44.4} & 25 \softscore{80.2} & 37.9 \softscore{55.0} \\
GPT-5.4-Mini + GPT-Audio-Mini & 47 \softscore{52.1} & 35 \softscore{40.5} & 50 \softscore{61.2} & 62 \softscore{82.5} & 23 \softscore{36.8} & 16 \softscore{22.2} & 60 \softscore{84.7} & 62 \softscore{72.7} & 36 \softscore{85.5} & 43.4 \softscore{59.8} \\
GPT-5.5 + GPT-Audio & \groupbest{74} \softscore{79.2} & \groupbest{73} \softscore{79.0} & \groupbest{77} \softscore{84.5} & \groupbest{66} \softscore{83.9} & 41 \softscore{53.1} & 29 \softscore{36.5} & \groupbest{77} \softscore{92.3} & \globalbest{\groupbest{83} \softscore{91.3}} & \groupbest{55} \softscore{94.5} & \groupbest{63.9} \softscore{77.2} \\

\midrule
\multicolumn{11}{l}{\textbf{Selected VLMs (vision only)}}\\[-2pt]
\midrule

Grok-4.3 & 55 \softscore{62.4} & 42 \softscore{50.4} & 52 \softscore{64.7} & \textemdash & \textemdash & \textemdash & 66 \softscore{85.9} & 57 \softscore{71.0} & 39 \softscore{88.7} & \textemdash \\
Gemma-4-31B & 36 \softscore{45.7} & 32 \softscore{42.9} & 47 \softscore{57.7} & \textemdash & \textemdash & \textemdash & 76 \softscore{92.3} & 70 \softscore{82.3} & 59 \softscore{95.0} & \textemdash \\
Qwen3.5-397B-A17B & 51 \softscore{56.5} & 40 \softscore{49.9} & 51 \softscore{63.3} & \textemdash & \textemdash & \textemdash & 78 \softscore{92.5} & 78 \softscore{87.8} & 35 \softscore{89.5} & \textemdash \\
Kimi-K3 & 55 \softscore{61.8} & 57 \softscore{64.9} & 60 \softscore{71.8} & \textemdash & \textemdash & \textemdash & 80 \softscore{92.6} & 79 \softscore{89.1} & 54 \softscore{93.8} & \textemdash \\
Claude-Fable-5 & \groupbest{76} \softscore{79.0} & \groupbest{62} \softscore{70.1} & \groupbest{85} \softscore{89.8} & \textemdash & \textemdash & \textemdash & \globalbest{\groupbest{90} \softscore{94.2}} & \groupbest{85} \softscore{91.2} & \groupbest{70} \softscore{96.0} & \textemdash \\

\bottomrule
\end{tabular}

\caption{
Model results on our benchmark comprising 900 tasks (100 per category). We report strict scores and soft scores (the smaller gray numbers in parentheses). The overall best score in each column is highlighted with a gray background and boldface, while the best score within each model group is underlined. The evaluation includes multimodal models, pairs of models from the same model family, and selected significant VLMs for comparison.
}

\label{tab:benchmark}
\end{table*}

\subsubsection{Verification rules for other tasks}
Apart from the categories focused on question answering, our benchmark also evaluates other competencies of multimodal models. The verification of other task types requires metrics tailored to the specific nature of each task. We have defined three additional verification rules for these tasks. Due to their higher complexity, they are only briefly outlined below. More detailed information regarding these metrics is provided in Appendix \ref{app:verification_rules}.

\noindent \textbf{Automatic speech recognition (ASR)} – Popular metrics for assessing speech transcription quality are the character and word error rates. In our benchmark, all metrics must be normalized to a range between 0 and 1. Therefore, we decided to measure word accuracy - the complement of the word error rate - defined as $\mathrm{WAcc} = \max(0, 1 - \mathrm{WER})$. The WAcc value is used directly as the soft score. Since achieving a perfect WAcc is challenging, for the strict score we adopted a cutoff threshold of 0.9, above which the task is considered successfully completed. If the response fails to exceed this threshold, the model receives 0 points for the task.

\noindent \textbf{Optical character recognition (OCR)} – Documents included in the benchmark may contain both text blocks and tables. In our approach, we employ separate metrics to evaluate the extraction accuracy of these elements. For text validation, we utilize the word accuracy (WAcc) metric, similar to speech recognition. If a document contains tables, each table is evaluated using the TEDS metric \citep{zhong2020image}, which measures both the structure and the content of the table. The overall score is calculated as a weighted average of WAcc score and all TEDS scores, where the weights are proportional to the number of characters of these elements in the extracted document. This value is used as the soft score. For the strict score, we applied a cutoff threshold of 0.9.

\noindent  \textbf{Structured extraction} – For this type of task, we expect the model to return a response in JSON format that complies with the JSON Schema provided as the input parameter. The response is validated by comparing the returned object with a reference object. For simplicity, all objects in our benchmark have either a flat structure or are arrays of objects with a flat structure. In other words, no object has nested objects, only fields of primitive types or arrays of primitive types. The task is considered correct if the model's response and the reference object contain identical values in their corresponding fields. The soft score is calculated as the ratio of matching field values to the total number of fields in both objects.

\section{Evaluation}
This section presents the evaluation results of multimodal models on our benchmark. We first report overall results, then compare general-purpose multimodal models with smaller specialized systems on ASR and OCR tasks.

The experiments include both open-weights models and commercial solutions accessible via API. To ensure results are as close as possible to deterministic, we initially set the temperature parameter to 0 for response generation. For reasoning models this sometimes caused a repetition loop. For each record where such repetition occurred, we iteratively increased the temperature by 0.3 until a correctly formatted response was obtained. All other generation parameters were kept at the default values suggested by the model's authors. Furthermore, some open-weights models, such as Gemma 4 \citep{farabet2026gemma4} and Qwen3.5 \citep{qwen3.5}, allow users to define a token budget for image representation. In such cases, we increased this budget to the maximum allowed values to preserve as much detail as possible in the model's input, which is particularly important for document-based categories.

\subsection{Main results}
The main results of our evaluation are presented in Table \ref{tab:benchmark}. It includes both models that support all the required modalities, as well as pairs of models, provided they originate from the same model family and are available in analogous variants e.g., GPT-4o and GPT-4o-Audio \citep{hurst2024gpt}, or Mistral-3.2-24B and Voxtral-24B \citep{mistralai2025mistralsmall31, liu2025voxtral}. At the time of our experiments, the number of multimodal models supporting the Polish language across text, image, and audio was relatively small. Since there are significantly more Visual Language Models (VLMs) available, we decided to also include popular examples of such models in the table for comparison purposes. A more comprehensive evaluation focusing solely on visual capabilities, covering more than 50 models, is provided in Appendix \ref{app:vlm_evaluation}.

\noindent \textbf{Frontier commercial models:} Among the most prominent commercial solutions, the Gemini models stand out. The Gemini-3.1-Pro \citep{google2025gemini3} model achieves the highest average score across all categories at nearly 80\%. The smaller versions of Gemini also offer good quality, the newer Gemini-3.5-Flash outperforms the Gemini-3.1-Pro in two document categories and one QA category. OpenAI does not offer a single model that supports all modalities, so our evaluation considers pairs of models, where one was responsible for visual categories and the other for audio categories. The best of the tested pairs scored around 64\%, falling short of the Gemini primarily in terms of audio understanding and extracting structure from documents. In knowledge-related categories, the lower performance of OpenAI models can be partially explained by refusals. GPT models refused to answer questions related to the recognition of figures or works of art, even historical ones, showing higher refusal rate than other model families. Claude-Fable-5 \citep{anthropic2026claude} is a strong model, delivering particularly good results in the document-based categories, where it performs on par with the Gemini models. However, due to its lack of audio support, we did not evaluate this model on the full benchmark.

\noindent \textbf{Open-weights models:} Phi-4-Multimodal-Instruct \citep{abouelenin2025phi} was one of the first open models to support both image and audio inputs. Unfortunately, despite its declared support for the Polish language, in practice it is insufficient given the difficulty level of the tasks. The model even struggled to formulate grammatically correct answers in Polish. Mistral models \citep{mistralai2025mistralsmall31,liu2025voxtral} perform slightly better, but still lag significantly behind commercial solutions. If we restrict our scope solely to vision tasks, much better open models are available. Specifically, Gemma 4 \citep{farabet2026gemma4} and Qwen3.5 \citep{qwen3.5} have brought significant improvements to image understanding, matching some commercial models in document-related categories despite their small size. Additionally, small Gemma 4 models in E2B and E4B variants support audio inputs, but limited to only 30 seconds, which is insufficient for our benchmark.

\noindent \textbf{Benchmark difficulty:} Sound \& music QA proved to be the most challenging category, with the highest score reaching only 56\%. This is understandable - model developers focus on other tasks with greater practical significance, making non-speech audio understanding less of a priority. The low scores in other categories can be attributed to our use of strict metrics, which require models to provide nearly perfect solutions. Structured extraction poses a problem for many models, as even a minor error in the JSON structure results in a failed response. This also explains the high variance in scores within this category. Models from the Qwen3.5 family lose the most ground here. While they demonstrate strong capabilities in other document-related categories, they struggle with extracting structured data from documents.

\begin{table}[ht]
  \centering
  \small
  \setlength{\tabcolsep}{3pt}
  \renewcommand{\arraystretch}{1.02}
  \begin{tabular}{l|c|c|c}
   \toprule
    \makebox[3.8cm]{} & \multicolumn{3}{c}{\textbf{Scores}} \\[-1pt]
    \hhline{~|---}
    \multirow{-2}{*}{\textbf{Model}} & \makebox[0.8cm]{Strict} & \makebox[1.0cm]{Soft} & \makebox[1.0cm]{Soft BN} \\[-1pt]
     \midrule
    \multicolumn{4}{l}{\textbf{ASR models}}\\[-2pt]
    \midrule
    Whisper-Tiny & 0 & 49.42 & 49.42 \\
    Qwen3-ASR-0.6B & 4 & 65.97 & 61.92 \\
    Whisper-Small & 22 & 75.82 & 75.82 \\
    Qwen3-ASR-1.7B & 24 & 74.79 & 72.26 \\
    Nvidia-Parakeet-TDT-0.6B-v3 & 36 & 79.80 & 78.87 \\
    Nvidia-Canary-1B-v2 & 38 & 78.88 & 77.58 \\
    Microsoft-VibeVoice-ASR & 41 & 80.27 & 79.13 \\
    Cohere-Transcribe-03-2026 & 46 & 81.51 & 80.29 \\
    Whisper-Medium & 51 & 82.68 & 82.68 \\
    Whisper-Large-V3-Turbo & 56 & 85.69 & 85.69 \\
    Whisper-Large-V3 & 63 & 86.40 & 86.40 \\
    GPT-4o-Transcribe & 66 & 86.45 & 85.67 \\
    ElevenLabs-Scribe-V2 & \globalbest{\groupbest{81}} & \globalbest{\groupbest{93.23}} & \groupbest{93.08} \\[-1pt]
    \midrule
    \multicolumn{4}{l}{\textbf{Audio-instruct models}}\\[-2pt]
    \midrule
    Voxtral-Small-24b-2507 & 36 & 80.27 & \textemdash  \\
    GPT-Audio-Mini & 62 & 82.55 & \textemdash \\
    GPT-4o-Audio & 64 & 82.57 & \textemdash \\
    GPT-Audio & \groupbest{66} & \groupbest{83.93} & \textemdash \\[-1pt]
    \midrule
    \multicolumn{4}{l}{\textbf{Multimodal models}}\\[-2pt]
    \midrule
    MiMo-V2-Omni & 3 & 48.56 & \textemdash  \\
    Gemini-3.1-Flash-Lite & 70 & 87.13 & \textemdash  \\
    Gemini-3.5-Flash & 77 & 89.20 & \textemdash  \\
    Gemini-3.1-Pro & 77 & 91.03 & \textemdash \\
    Gemini-2.5-Pro & \groupbest{81} & \groupbest{91.45} & \textemdash \\
    \bottomrule
  \end{tabular}
  \caption{ASR evaluation results.}
  \label{tab:asr}
\end{table}

\subsection{ASR and OCR evaluation}
The benchmark includes the automatic speech recognition (ASR) and optical character recognition (OCR) categories, which are based on well-established tasks of significant practical importance. These tasks attracted researchers' attention years before the emergence of large language models. Currently, there are also many specialized solutions focusing exclusively on these tasks. These solutions often rely on small models with parameter counts several orders of magnitude smaller than those of modern LLMs. Our goal was to use the challenging datasets we created for these two categories to compare small specialized models with larger multimodal models. Since most specialized models do not support instructions, we added an additional normalization step to adapt the output to our expected formatting, such as the format of numbers and tables. Details can be found in Appendix {\ref{app:verification_rules}}. \emph{Soft BN} column refers to raw soft scores before normalization.

\begin{table}[ht]
  \centering
  \small
  \setlength{\tabcolsep}{3pt}
  \renewcommand{\arraystretch}{1.02}
  \begin{tabular}{l|c|c|c}
    \toprule
    \makebox[3.8cm]{} & \multicolumn{3}{c}{\textbf{Scores}} \\[-1pt]
    \hhline{~|---}
    \multirow{-2}{*}{\textbf{Model}} & \makebox[0.8cm]{Strict} & \makebox[1.0cm]{Soft} & \makebox[1.0cm]{Soft BN} \\[-1pt]
     \midrule
    \multicolumn{4}{l}{\textbf{OCR models}}\\[-2pt]
    \midrule
    PaddleOCR-VL-1.5 & 8 & 47.05 & 44.87 \\
    GLM-OCR & 8 & 58.64 & 53.73\\
    Qianfan-OCR & 12 & 57.84 & 55.57 \\
    Tesseract-OCR & 13 & 35.13 & 32.07 \\
    Falcon-OCR & 19 & 57.52 & 55.34 \\
    HunyuanOCR & 21 & 52.05 & 48.36 \\
    DeepSeek-OCR-2 & 34 & 64.57 & 60.66 \\
    LightOnOCR-2-1B & 34 & 70.89 & 66.47 \\
    DotsOCR & 34 & 73.43 & 71.40 \\
    Nanonets-OCR2-3B & 35 & 74.55 & 72.05 \\
    RolmOCR & 38 & \groupbest{79.15} & \groupbest{79.05} \\
    Chandra-OCR-2 & \groupbest{45} & 78.54 & 64.78 \\[-1pt]
    \midrule
    \multicolumn{4}{l}{\textbf{Vision models}}\\[-2pt]
    \midrule
    GPT-4o & 17 & 31.78 & \textemdash \\
    Grok-4 & 17 & 59.54 & \textemdash \\
    GPT-5.4-Mini & 60 & 84.74 & \textemdash \\
    Kimi-K2.6 & 67 & 89.60 & \textemdash \\
    Gemma-4-31B & 76 & 92.34 & \textemdash \\
    GPT-5.5 & 77 & 92.32 & \textemdash \\
    Qwen3.5-397B-A17B & 78 & 92.52 & \textemdash \\
    Claude-Fable-5 &  \globalbest{\groupbest{90}} & \globalbest{\groupbest{94.20}} & \textemdash \\[-1pt]
    \midrule
    \multicolumn{4}{l}{\textbf{Multimodal models}}\\[-2pt]
    \midrule
    MiMo-V2.5 & 57 & 82.75 & \textemdash \\
    Gemini-3.1-Flash-Lite & 76 & 91.06 & \textemdash \\
    Gemini-2.5-Pro & 80 & 92.29 & \textemdash \\
    Gemini-3.1-Pro & 81 & 91.38 & \textemdash \\
    Gemini-3-Flash & \groupbest{86} & \groupbest{94.30} & \textemdash \\[-1pt]
    \bottomrule
  \end{tabular}
  \caption{OCR evaluation results.}
  \label{tab:ocr}
\end{table}

\noindent \textbf{ASR results:} Table \ref{tab:asr} shows the results on the ASR task. The best result was achieved by the commercial model from ElevenLabs, but Gemini 2.5 \citep{comanici2025gemini} and 3.1 \citep{google2025gemini3} achieved similar results. Among the open models, the Whisper family \citep{radford2023robust} performs surprisingly well in Polish. Whisper-Large-V3 proved to be only slightly worse than the closed services offered by OpenAI: GPT-4o-Transcribe and GPT-Audio. Newer open-weight models such as Qwen3-ASR \citep{shi2026qwen3}, VibeVoice \citep{peng2026vibevoice}, Cohere-Transcribe \citep{cohere2026transcribe}, and Nvidia models \citep{sekoyan2025canary} also fall behind Whisper-Large-V3, which was released in 2023. These results differ from those reported on the Open ASR Leaderboard \citep{srivastav2025open}, which focuses mainly on English and, to a lesser extent, on a few of the most popular European languages. This may suggest that in the speech recognition task, performance on high-resource languages does not necessarily correlate with performance on less popular ones.

\noindent \textbf{OCR results:} Table \ref{tab:ocr} summarizes the OCR task results. The Claude-Fable-5 \citep{anthropic2026claude} model achieved the best result in this task. Among the multimodal models, the Gemini family of models achieved high scores, particularly the Gemini-3.5-Flash \citep{google2025gemini3} model. Models dedicated to OCR achieved significantly lower scores under the strict metric. The best model in this group was Chandra-OCR-2\footnote{\url{https://hf.co/datalab-to/chandra-ocr-2}}. At the same time, looking at the soft metric results, aside from the Chandra-OCR-2 model, four other models from this group, namely RolmOCR \citep{RolmOCR}, Nanonets-OCR2-3B \citep{Nanonets-OCR2}, DotsOCR \citep{Li2025dotsocrMD}, and LightOnOCR-2-1B \citep{lightonocr2_2026}, achieved scores above 70\%, indicating that a majority of the extracted text was correct. Some of the evaluated OCR-oriented models did not respond correctly to the prompt for this task. To mitigate the negative impact of this behavior, we implemented the normalization described in the Appendix \ref{app:verification_rules_ocr}. 

\section{Conclusions}
We introduced PUMA, a new benchmark for evaluating multimodal language models in the Polish linguistic and cultural context. The benchmark consists of manually created tasks spanning images, audio, and documents, and covers both culturally grounded question answering and practical skills. By relying on deterministic rule-based verification, it enables reproducible and cost-efficient evaluation without external judges. Our evaluation shows that current multimodal models still differ substantially in their ability to handle Polish multimodal inputs. Frontier commercial models achieve strong results, especially in visual question answering and document understanding, but important gaps remain. Audio understanding, particularly non-speech sound and music interpretation, is still challenging for all evaluated systems. These findings suggest that progress in multimodal AI cannot be assessed reliably using English-centric or text-only benchmarks alone. Local language, cultural grounding, and realistic document and audio inputs reveal limitations that are otherwise easy to miss.

\section*{Limitations}

\paragraph{Rule-based evaluation.}
The benchmark uses a rule-based evaluation approach, which requires questions to be designed in a structured way with clearly defined answers and verification rules. This limits the types of tasks that can be included, especially highly open-ended ones. At the same time, this approach enables fast, low-cost, and consistent evaluation without the need for human judges or additional language models. It also makes the evaluation largely deterministic, which improves reproducibility. Rule-based evaluation covers question answering categories. The remaining categories - ASR, OCR, and Structured Extraction - are evaluated using task-specific metrics commonly applied to these types of problems.

\paragraph{Modality coverage.}
The benchmark does not include a video modality. This was a deliberate design choice, as support for video understanding is still limited in current multimodal models, which would reduce the number of models that could be evaluated. However, the annotation and evaluation framework supports video data, allowing future extensions of the benchmark.

\paragraph{Temporal coverage.}
Most questions focus on well-known facts, historical events, and widely recognized cultural elements. While some tasks, particularly in the "Contemporary Life" category, include more recent knowledge (up to 2025), we avoided questions that depend on rapidly changing information or short-lived trends. This improves long-term stability but limits coverage of very recent phenomena.

\paragraph{Cultural diversity.}
The benchmark was designed to cover a wide range of cultural domains through manual curation and careful selection of tasks. Annotators actively introduced diverse topics across categories to avoid strong underrepresentation of specific areas. While defining and measuring cultural diversity remains challenging, the current approach provides a strong and flexible foundation that can be further extended in future work.

\paragraph{Annotation process and consistency.}
The annotator group was relatively homogeneous in terms of age and experience, which may limit coverage of perspectives specific to other generations. At the same time, this helped ensure consistency in annotation and a shared understanding of cultural context. In addition, annotators had comparable experience in working with language models and evaluation tasks, which supported the design of clear and robust verification rules. PUMA annotation involved creating complete multimodal tasks, including instructions, expected answers, and verification rules, rather than assigning nominal labels to a shared set of examples. Consequently, standard agreement measures such as Cohen's $\kappa$ or Krippendorff's $\alpha$ would be difficult to apply and interpret. Instead, we used a consensus-based quality control process in which three annotators reviewed questions and verification criteria according to predefined acceptance rules and discussed identified issues. The process was inspired by other established benchmarks that employed similar approaches \citep{hendrycks2026benchmark,wang2024mmlu,wei2024measuring}. This procedure improved dataset consistency, but it does not provide quantitative evidence that independent annotators would construct comparable tasks.

\section*{Ethical considerations}

\paragraph{Privacy and data sources.}
Most media files used in the benchmark come from publicly available sources. A small number of files were created by the authors or come from their own private collections. The questions themselves were written manually by the annotation team and are original. When personal data appears in the dataset, it refers only to well-known public figures.

\paragraph{Benchmark usage and data release.}
To reduce the risk of data leakage, only a subset of the dataset will be publicly released. The full benchmark will remain private. At the same time, we plan to provide a public leaderboard with evaluation results, which will be regularly updated with new models. This approach aims to balance transparency with the integrity of the benchmark.

\paragraph{Use of AI tools.}
AI tools were used in the preparation of the publication for only language editing, grammatical correction and minor stylistic improvements. They are not used for generating the core scientific content or for creating the benchmark itself.

\section*{Acknowledgments}
The research was supported by the project Large Language Models for the European Union (LLMs4EU). This project is co-funded by the Digital Europe Programme under Grant Agreement 101198470.

\bibliography{custom}

\clearpage
\appendix

\renewcommand*\footnoterule{} 
\newcommand\blfootnote[1]{
  \begingroup
  \renewcommand\thefootnote{}\footnote{#1}
  \addtocounter{footnote}{-1}
  \endgroup
}

\section{Verification rules in detail}
\label{app:verification_rules}
This section describes additional details explaining the evaluation in three special categories: automatic speech recognition (ASR), optical character recognition (OCR), and structured extraction. Unlike in the question answering categories, where the prompt for the model consists of the question or command itself, for these three categories we apply a single universal prompt across the entire category. This prompt contains a task description and guidelines regarding the format in which the answer should be returned.

\subsection{Automatic speech recognition (ASR)}
To evaluate the quality of speech recognition, we use the WAcc metric, which is the complement of the word error rate, additionally limited to the 0-1 range: $\mathrm{WAcc} = \max(0, 1 - \mathrm{WER})$. For the strict score, the model must achieve at least 0.9 for the question to be considered passed. For the soft score, the WAcc value is used directly. The comparison between the reference text and the model's output is performed on normalized texts. When calculating the metric, only words (alphanumeric strings) are taken into account, excluding other character types such as punctuation, whitespace, or control characters. Additionally, word comparison is case-insensitive.

If the evaluated model supports instructions, we provide it with the following prompt:
\begin{tcolorbox}[enhanced jigsaw,breakable,pad at break*=1mm,colback=blue!5!white,colframe=blue!50!white,title=Default prompt for ASR]
  \small
  \textbf{Original prompt:} Dokonaj transkrypcji pliku audio w języku polskim. Wszystkie wartości liczbowe powinny być zapisywane pełnymi słowami i bez stosowania skrótów np. "sto dwadzieścia pięć złotych" zamiast "125 zł". Zwróć tylko zawartość nagrania, bez dodatkowych komentarzy.
  \tcblower
  \small
  \textbf{English translation:} Transcribe this audio file in Polish. All numerical values should be written out in full words and without using abbreviations, for example, "one hundred twenty-five zlotys" instead of "125 zł". Return only the content of the recording, without any additional comments.
\end{tcolorbox}

\textbf{Numbers normalization} - All numeric values in the reference transcripts are written out in full words. The above prompt forces the models to return numbers in the same format to enable the correct calculation of the WAcc metric. However, some of our experiments also included different types of models trained exclusively on the speech recognition task. Such models do not support instructions, so we have no control over how numbers are formatted in the output. To evaluate these models, we applied an additional normalizer based on the \emph{Gemini 3 Flash} model. If the model's response contained digits, we post-processed this response by sending it to the normalizer with the following prompt:

\begin{tcolorbox}[enhanced jigsaw,breakable,pad at break*=1mm,colback=blue!5!white,colframe=blue!50!white,title=Normalization prompt]
  \small
  \textbf{Original prompt:} Zmodyfikuj poniższy tekst, zamieniając w nim wszystkie liczby na pełny zapis słowny (np. '125 zł' -> 'sto dwadzieścia pięć złotych', '12:39' -> 'dwunasta trzydzieści dziewięć', '2004 r.' -> 'dwa tysiące czwarty rok'). Zwróć tylko poprawiony tekst, nie dodawaj żadnych komentarzy:
  \tcblower
  \small
  \textbf{English translation:} Modify the text below by replacing all numbers in it with their full word form (e.g., '125 zł' -> 'one hundred twenty-five zlotys', '12:39' -> 'twelve thirty-nine', '2004 r.' -> 'two thousand fourth year'). Return only the corrected text, do not add any comments:
\end{tcolorbox}

\subsection{Optical Character Recognition (OCR)}
\label{app:verification_rules_ocr}

To evaluate OCR quality, we use a hybrid metric that combines textual accuracy with table structure correctness. All non-tabular elements are reduced to plain text, while tables are evaluated as structured content.

For the textual component, we follow the same approach as in ASR and use the WAcc metric, defined as the complement of the word error rate, bounded to the range [0,1]. The comparison between the reference text and the model output is performed on normalized text. Only words (alphanumeric tokens) are considered, while punctuation, whitespace, and other non-alphanumeric characters are ignored. The comparison is also case-insensitive.

For table similarity, we employ the Tree-Edit-Distance-based Similarity (TEDS) metric introduced by \citet{zhong2020image}. TEDS measures the similarity between predicted and ground-truth tables by representing them as ordered HTML trees and computing their normalized tree edit distance. In this representation, the HTML table content is structured such that its children are table rows (\texttt{tr}), while the leaf nodes correspond to table cells (\texttt{td}). Each cell node is associated with three attributes: \texttt{colspan}, \texttt{rowspan}, and textual \texttt{content}.

The similarity between two tables is computed using the tree edit distance formulation of \citet{pawlik2016apted}, where the cost of insertion and deletion operations is set to 1. For node substitution, the cost depends on node types and attributes. Specifically, substituting two nodes incurs a cost of 1 if at least one of them is not a \texttt{td} node. When both nodes correspond to table cells, the substitution cost is 1 if their structural attributes (i.e., \texttt{colspan} or \texttt{rowspan}) differ. Otherwise, the substitution cost is defined as the normalized Levenshtein similarity between their textual contents, allowing partial credit for OCR outputs that are textually similar but not identical.

Finally, TEDS between two trees is computed as:
\begin{equation}
\mathrm{TEDS}(T_a, T_b) = 1 - \frac{\mathrm{EditDist}(T_a, T_b)}{\max(|T_a|, |T_b|)}
\end{equation}

where $\mathrm{EditDist}(T_a, T_b)$ denotes the minimum-cost sequence of edit operations transforming the predicted tree $T_a$ into the ground-truth tree $T_b$, and $|T|$ is the number of nodes in a tree.

To account for ambiguities in reading order and segmentation, the reference document is decomposed into coherent text blocks in cases where the layout contains regions with non-unique or ambiguous reading order. Such blocks are defined during dataset construction when multiple valid linearizations of the content are possible. Since OCR systems may produce content in different orders, we consider permutations of the reference blocks. For each permutation, the concatenated reference text is compared against the full OCR output, and the highest score is retained. To ensure computational tractability, the number of blocks is limited to at most 4, which bounds the number of evaluated permutations. This makes the evaluation robust to variations in layout reconstruction and reading order while remaining computationally feasible.

For tabular content, we adopt a matching strategy that is independent of table ordering. Each reference table is compared against all tables extracted from the model output, and the highest TEDS score is selected. The final table similarity score is computed as a weighted average of TEDS scores over all reference tables, where the weight of each table is proportional to its size, defined as the ratio of its length to the length of the largest reference table. If the model produces fewer tables than present in the reference, missing tables are assigned a score of 0 with their corresponding weights included in the average. This design penalizes incomplete table detection while remaining invariant to table ordering.

The final score is computed as a weighted average of WAcc (for text) and the table similarity score, where the weights depend on the relative proportion of textual and tabular content in the reference. For the strict score, the model must achieve at least 0.9 to be considered correct, while for the soft score, the metric value is used directly. If multiple valid reference segments are available, all permutations are evaluated and the best score is selected.

If the evaluated model supports instructions, we provide it with the following prompt:

\begin{tcolorbox}[enhanced jigsaw,breakable,pad at break*=1mm,colback=blue!5!white,colframe=blue!50!white,title=Default prompt for OCR]
    \small
    \textbf{Original prompt:} Przekonwertuj poniższy dokument w języku polskim do formatu markdown. Zwróć tylko markdown bez dodatkowych komentarzy. \\
    ZASADY: \\
    - Zwróć tekst zgodnie z kierunkiem jego czytania: od góry do dołu i od lewej do prawej. \\
    - Słowa przeniesione z myślnikiem do nowej linii połącz w całość. \\
    - Musisz uwzględnić wszystkie tekstowe informacje znajdujące się na stronie. Nie pomijaj nagłówków, stopek, tabel, podpisów pod ilustracjami, przypisów. \\
    - Pomiń numerację stron. \\
    - Pomiń zdjęcia, ilustracje, znaki wodne, pieczątki i inne elementy graficzne. Nie dodawaj żadnych komentarzy dotyczących elementów graficznych. \\
    - Tabele zapisz w formacie HTML. \\
    - Preferuj używanie znaków \checkbox{0} i \checkbox{1} do prezentacji check boxów.
    \tcblower
    \small
    \textbf{English translation:} Convert the following Polish document into Markdown format. Return only the Markdown without any additional comments. \\
    RULES: \\
    - Return the text in reading order: from top to bottom and from left to right. \\
    - Merge words split with a hyphen across lines. \\
    - Include all textual information present on the page. Do not omit headers, footers, tables, figure captions, or footnotes. \\
    - Ignore page numbering. \\
    - Ignore images, illustrations, watermarks, stamps, and other graphical elements. Do not add any comments about graphical elements. \\
    - Represent tables using HTML format. \\
    - Prefer using \checkbox{0} and \checkbox{1} symbols for checkboxes.
\end{tcolorbox}

In addition to general-purpose LLMs, we also evaluate OCR-oriented methods, including approaches based on smaller fine-tuned language models as well as partially rule-based systems. Since such methods often produce outputs with method-specific formatting (e.g., auxiliary tags or non-standard structural markers), we apply a lightweight post-processing step to normalize their outputs to a unified representation of plain text and HTML tables.

This post-processing is performed using a large language model (Gemini 3 Flash), similarly to the normalization procedure applied in ASR. Importantly, the model is constrained to perform only structural normalization, without altering the linguistic content. In particular, it is instructed not to correct spelling, modify words, change punctuation or casing, or paraphrase the text. The allowed operations are limited to removing repetitive trailing fragments, merging hyphenated words split across lines, removing non-textual elements (e.g., image or watermark tags), and reconstructing spaced letter sequences into words.

The post-processing prompt used for OCR normalization is provided below.

\begin{tcolorbox}[enhanced,breakable,pad at break*=1mm,colback=blue!5!white,colframe=blue!50!white,title=Normalization prompt]
    \small
    \textbf{Original prompt:} Twoim zadaniem jest wykonanie WYŁĄCZNIE operacji normalizacji strukturalnej tekstu OCR. \\
    
    \textbf{ZASADY OGÓLNE:} \\
    - NIE poprawiaj literówek. \\
    - NIE zmieniaj słów, nawet jeśli są błędne. \\
    - NIE zmieniaj interpunkcji. \\
    - NIE zmieniaj wielkości liter. \\
    - NIE parafrazuj tekstu. \\
    - NIE zmieniaj polskich słów na angielskie. \\
    - NIE zmieniaj ostatniego słowa w tekście, nawet jeśli jest to niepełny wyraz z myślnikiem. \\
    - NIE dodawaj ani nie usuwaj treści poza przypadkami opisanymi poniżej. \\
    - Traktuj tekst jako surowe dane, nie jako język naturalny do poprawy. \\
    - WAŻNE: Jeśli nie masz pewności czy coś należy zmienić -- NIE zmieniaj tego. \\

    \textbf{DOZWOLONE OPERACJE (i tylko te):} \\
    1. Jeśli koniec tekstu składa się z wielokrotnych powtórzeń jednego wyrazu lub frazy, a tych powtórzeń jest co najmniej trzy -- usuń powtórzenia zostawiając tylko jedno wystąpienie tego wyrazu / frazy. \\
    2. Jeśli słowo jest rozdzielone myślnikiem na końcu linii -- zapisz je jako jeden wyraz. \\
    3. Usuń opisy elementów graficznych (np. Obraz:, Rysunek:, Zdjęcie:, Pieczątka:, Watermark:, a także tagi, np. [image], <image>). \\
    4. Połącz sekwencje pojedynczych liter w wyrazy oddzielone spacją. \\

    \textbf{PRZYKŁADY:} \\
    <<Przykłady usunięte dla czytelności>> \\
    
    \textbf{WAŻNE:} \\
    Każda zmiana musi wynikać BEZPOŚREDNIO z powyższych reguł. W przeciwnym razie tekst musi pozostać identyczny znak po znaku. \\
    Zwróć wyłącznie przetworzony tekst, bez komentarzy.
    \tcblower
    \small
    \textbf{English translation:} Your task is to perform \textbf{only} structural normalization of OCR text. \\

    \textbf{GENERAL RULES:} \\
    - Do not correct spelling errors. \\
    - Do not change words, even if they are incorrect. \\
    - Do not modify punctuation. \\
    - Do not change letter casing. \\
    - Do not paraphrase the text. \\
    - Do not translate Polish words into English. \\
    - Do not modify the last word, even if it is incomplete due to hyphenation. \\
    - Do not add or remove content except as specified below. \\
    - Treat the text as raw data, not natural language to improve. \\
    - IMPORTANT: If you are unsure whether a change is needed, do not change it. \\

    \textbf{ALLOWED OPERATIONS (only these):} \\
    1. If the end of the text consists of repeated occurrences of the same word or phrase (at least three times), remove repetitions and keep only one instance. \\
    2. If a word is split with a hyphen at the end of a line, merge it into a single word. \\
    3. Remove descriptions of graphical elements (e.g., Image:, Figure:, Photo:, Stamp:, Watermark:) and tags (e.g., [image], <image>). \\
    4. Merge sequences of space-separated characters into words. \\

    \textbf{EXAMPLES:} \\
    <<Examples omitted for clarity>> \\

    \textbf{IMPORTANT:} \\
    Every modification must follow directly from the rules above. Otherwise, the text must remain identical character by character. \\
    Return only the processed text, without any comments.
\end{tcolorbox}

\subsection{Structured extraction}
The evaluation of the structured extraction task involves comparing two JSON structures: the ground truth and the one returned as the model's response. When creating a new record, the annotator was tasked with defining a JSON schema describing the expected response format, preparing a reference object representing the correct answer, attaching the document image, and adding model instructions detailing the specific extraction task for that document. Regardless of these detailed instructions, a \emph{response\_format} parameter is included in every request sent to the model, passing the defined JSON schema. Most providers and popular open-source LLM serving frameworks (such as vLLM or SGLang) support structured outputs via \emph{response\_format}, which constrains the model to generate a response that conforms to the expected structure. For models that do not support this functionality, we also append a universal prompt to each request with the following content:

\begin{tcolorbox}[enhanced jigsaw,breakable,pad at break*=1mm,colback=blue!5!white,colframe=blue!50!white,title=Default prompt for structured extraction]
  \small
  \textbf{Original prompt:} Wyodrębnij informacje z załączonego dokumentu i zwróć odpowiedź w formacie JSON zgodnym z nastepującym JSON schema: \\
  \texttt{\{JSON\_SCHEMA\}}
  \tcblower
  \small
  \textbf{English translation:} Extract the information from the attached document and return the response in JSON format according to the following JSON schema: \\
  \texttt{\{JSON\_SCHEMA\}}
\end{tcolorbox}

The benchmark imposes certain restrictions on the complexity of structures that can be defined within this category. Only objects with a flat structure are allowed, meaning they must contain fields of primitive types or fields containing arrays of primitive types. It is also possible to define the ground truth as an array of such objects. If the model's response is not a valid JSON, it receives 0 points. If the JSON is valid, the ground truth is compared with the structure returned by the model.

\textbf{Object comparison} – The similarity between a pair of objects is measured as the number of fields for which both objects have the same values, relative to the total number of fields in those objects. More formally, if $F_R$ is the set of fields in the reference object and $F_A$ is the set of fields in the response object, and $v_R(k)$, $v_A(k)$ are functions returning the value for field $k$ in the reference and response objects respectively, then the similarity between the two objects is given by the formula:
\begin{equation}
    S(F_R, F_A) = \frac{\left|\left\{k \in F_R \;:\; v_R(k) = v_A(k)\right\}\right|}{\left| F_R \cup F_A \right|}
\end{equation}

\noindent Numeric and boolean values must match exactly to be considered equal. In the case of string values and arrays of strings, they are normalized before comparison by removing all non-alphanumeric characters and converting them to lowercase. When comparing text, only word consistency matters, ignoring punctuation, whitespace, and case sensitivity. The similarity value $S(F_R, F_A)$ is used directly as the soft score. The strict score is equal to 1 only if both objects are identical, i.e., when the similarity value is 1.

\textbf{Array comparison} – Ground truth can also be an array of objects. When comparing a pair of arrays, the order of elements does not matter. The similarity is calculated by comparing objects in pairs, and the total similarity is the average of the similarities between all matched pairs of objects, where unmatched objects receive a similarity of 0. The validation algorithm finds the optimal alignment of objects in both arrays - the one that maximizes the total similarity value.

\textbf{Array comparison for OpenAI models} - One limitation of the official OpenAI API at the time of our experiments was the lack of support for defining arrays as top level JSON Schema values passed in the \emph{response\_format} parameter. For these models, we use a workaround that involves wrapping the array in an additional object and then extracting the array from that object before performing the validation.

\section{Image categories}
\label{app:image_categories}
These categories focus primarily on verifying the models' knowledge of the Polish cultural context. Each example consists of a single image file and a prompt containing a question and an instruction for the model. If a question requires reasoning across multiple images, we combine them into a collage and save it as a single image. The instruction usually specifies the required output format, which simultaneously evaluates the models' instruction-following capabilities and facilitates rule-based parsing of the answers. Below are the annotator instructions for each category, along with the tag distribution for the questions created within the benchmark.

\subsection{History and culture}
\textbf{Instruction for annotators} - This category covers topics related to the history, cultural heritage, traditions, and customs of Poland. The visual materials included in it may present architectural landmarks, works of classical art (including sculpture and painting), and artifacts related to key events in Polish history. Furthermore, this category encompasses products of material culture, including everyday objects, traditional attire, machinery, built infrastructure, and sacral buildings. The use of photographs depicting specific figures, objects, and geographical locations is permitted, provided they have historical or cultural significance and the question itself refers directly to this context. However, materials depicting books and other written texts whose analysis requires character recognition are excluded from the scope of this category (such sources, regardless of their historical value, will be placed in document categories). Exceptions to this rule are general shots of libraries and historical book covers. Culinary-related items (food products, dishes) may be included in this category only if the subject matter of the question is embedded in a historical context, explicitly excluding strictly contemporary photos.

\noindent \textbf{Distribution of tags in the category} - tradition (11), architecture (9), world war II (9), monuments (8), cuisine (8), visual arts (6), prl (5), television (5), literature (5), historic event (4), middle ages (4), music (3), film (3), items (3), kings (3), religion (3), crafts (3), maps (3), world war I (3), heraldry (3), prehistoric (2), clothes (2), poles (2), science (2), sports (2), place (2), theater (2), performing arts (1), other (1), modernity (1), popculture (1), antiques (1), polish-soviet war (1), people (1), furniture (1), brands (1)

\subsection{Contemporary life}

\textbf{Instruction for annotators} - This category focuses on issues related to contemporary culture and the realities of everyday life in Poland. Its thematic scope covers broadly understood mass culture (including cinematography, television productions, radio, theater, and comics) as well as significant events, including official ceremonies, concerts, and mass gatherings. In addition, the category includes phenomena from the realm of digital culture, such as memes, symbols widely recognized in Poland, logotypes, and popular internet platforms. Another important element is the visual aspects of everyday Polish life – popular product brands or characteristic social habits and behaviors with a strong visual component. The scope of this category is complemented by Polish cuisine. If a question refers to a specific event, it should be noted that we consider contemporary events to be those that took place after 1989.

\noindent \textbf{Distribution of tags in the category} - characters (22), places (14), television (13), art (8), memes (7), food (7), brands (6), architecture (6), shows (6), music (5), politics (5), sports (5), film (5), virals (3), cartoon (3), events (3), goods (3), tradition (3), prl (3), games (3), stereotypes (2), cinema (2), logotypes (2), sayings (2), entertainment (2), nature (2), products (2), sculptures (2), cars (2), clothes (2), historic events (2), warsaw (2), monuments (2), comedy (2), comics (1), elections (1), crafts (1), internet (1), technology (1), leisure (1), inventors (1), charity (1), place (1), zoo (1), animation (1), modern (1), historical (1)

\subsection{Geography and environment}

\textbf{Instruction for annotators} - This category verifies the models' knowledge of natural and anthropogenic features located within the territory of Poland, as well as native flora and fauna. Regarding human-made features, the photos may present diverse settlement areas (urban and rural landscapes), architecture, and characteristic landmarks. Natural features, on the other hand, include various landforms and ecosystems, such as forests, lakes, islands, mountain peaks and ranges, desert areas, and coastal regions. The visual representation of Poland's geographical conditions may take the form of direct landscape photographs or various types of maps. An important aspect of this category is biology and ecology. Tasks include identifying species of animals, plants, and fungi found in Poland, as well as recognizing their specific parts (e.g., leaves, fruits, vegetables). Furthermore, the category incorporates topics from the fields of climatology and geology, based on illustrations of meteorological phenomena, soil types, rock formations, minerals, and deposits of natural resources. The category also includes questions regarding human interference in the natural environment, including large-scale infrastructure such as mines, water dams, tunnels, bridges, and shipping canals.

\noindent \textbf{Distribution of tags in the category} - places (25), maps (14), plants (13), mountains (11), animals (11), cities (10), rocks (9), National Parks (8), infrastructure (6), lakes (5), birds (5), forests (4), villages (3), rivers (2), beaches (2), mushroom (2), fruits (2), symbols (1), soil (1), waterfalls (1)

\section{Audio categories}
\label{app:audio_categories}

\begin{figure}
  \centering
  \includegraphics[scale=0.7]{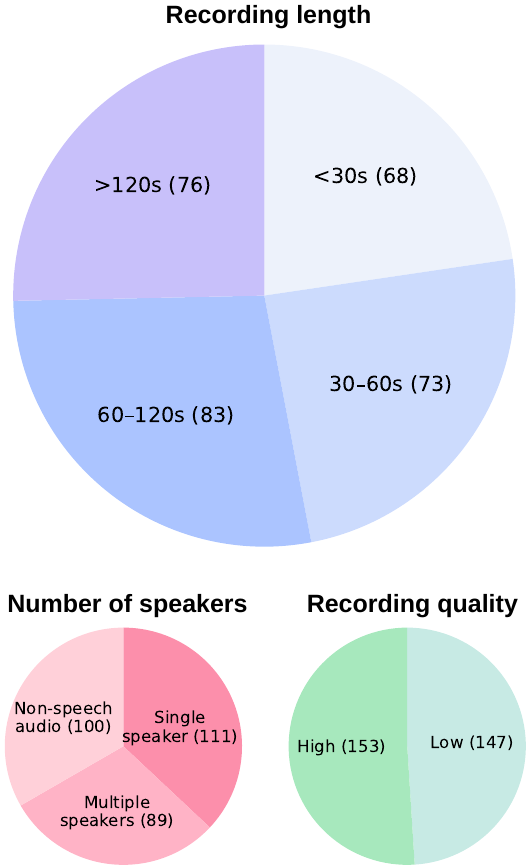}
  \caption{The number of examples in audio categories (ASR, Speech QA, Sound and music QA) by recording length, number of speakers and recording quality.}
  \label{fig:audio_chart}
\end{figure}

Two of the categories test the models' knowledge and reasoning abilities based on audio materials, similarly to the image categories. Meanwhile, one category evaluates the models' ability to transcribe speech to text (ASR). For the knowledge-based categories, each example consists of a single audio file, ranging from a few seconds to a few minutes in length, and a prompt for the model. Across all three categories, the mean recording duration is 82 seconds: 119 seconds for ASR, 73 seconds for Speech QA, and 54 seconds for Sound \& Music QA. To better illustrate the distribution of audio samples in our benchmark, Figure \ref{fig:audio_chart} shows the number of samples broken down by duration, number of speakers, and recording quality. For the ASR category, the example consists solely of a speech recording, while we use the same prompt for all examples. Below are the annotator instructions for each category, along with the tag distribution for the questions created within the benchmark.

The ASR category was designed to evaluate model performance under challenging, realistic conditions. We therefore selected more demanding recordings involving factors such as background noise, multiple speakers, dialectal speech, unclear diction, and lower recording quality. Below, we provide the annotator instructions for each category and the distribution of tags among the benchmark questions.

\begin{figure*}
  \centering
  \includegraphics[scale=0.85]{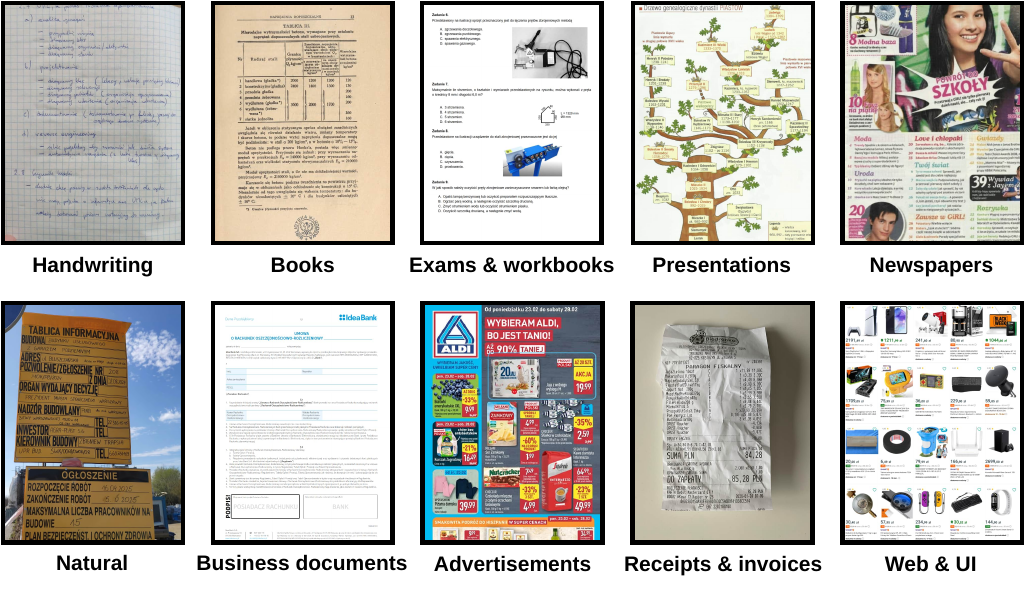}
  \caption{The types of documents included in our benchmark with examples.}
  \label{fig:document_types}
\end{figure*}

\subsection{Automatic speech recognition}
\textbf{Instruction for annotators} - This is a specific category related to converting speech to text. This category should contain short audio excerpts with a preferred length ranging from about a minute to a maximum of a few minutes, containing the utterances of one or more people. Audio samples for ASR can be divided into two groups based on their quality. The first group consists of studio-quality samples, i.e., recorded in a studio environment using professional equipment, where the speakers' utterances are clear and undisturbed by external factors. These can be excerpts from audiobooks, podcasts, movies, cartoons, or television programs. Such recordings may also contain other sounds – jingles, background music, audience noises – which are intentionally introduced elements of the recording. The second group consists of amateur-quality samples, characterized by various types of interference resulting from the recording equipment, unclear speech, unintentional background noises, or multiple people speaking simultaneously. This group includes telephone and internet conversations, street interviews, lectures, and speeches recorded with amateur equipment from a long distance from the speaker, as well as any utterances disrupted by ambient noises. Recordings where the speaker's diction poses a problem (e.g., emotions, a raised voice, shouting, unusual word accentuation) can also be a challenge for the models. Instruction for the model should not be added, the same universal prompt is used for all examples in this category.

\noindent \textbf{Distribution of tags in the category} - low quality (67), single speaker (54), multi speaker (46), high quality (33), television (20), lecture (15), interview (12), speech (12), dialog (11), historic event (8), shows (6), music (5), internet (5), book (4), language (1), politics (1)

\subsection{Speech question answering}

\textbf{Instruction for annotators} - This category focuses on evaluating the ability of language models to understand and interpret audio recordings in Polish. Tasks within this section are divided into two groups: strictly linguistic questions, where the correct answer stems solely from the analysis of the recording's content, and questions requiring the utterance to be embedded in a broader historical and cultural context. This latter type of task expects the model not only to correctly understand the speech but also to integrate it with its preexisting knowledge about Poland.

The first group includes questions taking into account the Polish cultural and historical context, requiring the combination of audio material analysis with general knowledge. This set includes tasks based on quote attribution, i.e., linking famous statements with the appropriate historical background or the events they concerned. Furthermore, this category entails the identification of widely known historical and public figures based on their voice and speaking characteristics. Models are also expected to demonstrate broad contextual understanding, consisting of interpreting the content of a statement in relation to knowledge about Poland and its traditions, as well as recognizing works of pop culture, such as Polish movies, TV series, or radio plays, based on short audio snippets.

The second group of tasks verifies the understanding of Polish speech and is based on the analysis of the recording itself. It should include questions testing both general and detailed comprehension of the content and intention of the utterance. The difficulty level of the question can be increased by using recordings with difficult language, for example, containing industry-specific vocabulary, slang, dialects, or vernaculars. Tasks in this area may involve the extraction of specific information (such as proper nouns or names), recognizing the speaker's intention, interpreting a joke, as well as analyzing the argumentation used by the parties in a dialogue. The model's ability to assess the emotional state of the interlocutors, deduce the causes of this state, and determine the interpersonal relationships connecting the heard individuals is also subject to verification. Another aspect within this group is the analysis of the audio structure. This includes evaluating dialogue parameters, such as counting speakers, estimating the proportion of speaking time for individual people, or identifying the dialect used. The category may also include tasks related to verifying linguistic correctness, consisting of finding and correcting errors present in the recorded utterances.

\noindent \textbf{Distribution of tags in the category} - high quality (57), single speaker (57), low quality (43), multi speaker (43), reasoning (33), people (31), television (14), historic event (8), place (7), dialect (6), language (6), music (5), events (3), book (2), theater (1), radio (1), emotion (1), opera (1), dance (1)

\subsection{Sound and music question answering}
\textbf{Instruction for annotators} - This category covers tasks based on the analysis of non-speech audio materials that are tied to Poland or the Polish cultural context. The scope of this category includes, among others, musical tracks, the sounds of instruments and tools, animal sounds, as well as ambient noises associated with characteristic locations that can be recognized by sound. Additionally, it incorporates audio signals and jingles related to Polish institutions, brands, or television programs.
The methodology allows the use of sound samples not explicitly related to Poland, provided that the content of the question unambiguously refers to the Polish context (e.g., a presentation of a piano sound combined with a question about prominent Polish pianists). Materials assigned to this category may contain trace elements of human speech; however, their analysis and comprehension cannot be a prerequisite for providing the correct answer. Any tasks requiring speech understanding should be classified under the Speech QA category.

\noindent \textbf{Distribution of tags in the category} - high quality (63), low quality (37), music (30), people (21), television (19), instrument (13), urban (9), events (9), jingle (8), place (7), animals (6), folk (4), nature (3), historic event (3), dance (2), music theory (2), opera (1)

\section{Document categories}
\label{app:document_categories}
Document categories evaluate the capabilities of models in understanding documents provided in a graphical format, along with all the elements that may appear within them, such as text blocks, tables, charts, photos, infographics, diagrams, or handwritten notes. In these categories, similarly to the image categories, we provide the model with a file representing a fragment of a document - usually a single page - as well as an instruction or question in the form of a prompt. In the OCR category, the instruction is the same for all examples, whereas in the remaining categories, each example has an individual prompt.

Unlike the other categories, we have used a formalized tagging structure for the document categories. Each example should include three tags specifying the type of document, its date of creation, and its quality. For the purposes of our work, we have identified the following document types:
\begin{itemize}[wide,labelwidth=0pt,labelindent=0pt,itemsep=0pt,topsep=5pt,parsep=0pt]
\item \textbf{Handwriting} – notes, letters, postcards, and any other handwritten documents.
\item \textbf{Books} – books, articles, and documents containing primarily continuous text and tables, with a small number of graphical elements.
\item \textbf{Exams \& workbooks} – examination papers and books containing tasks or exercises, where, alongside a small amount of text, there are fields to fill in, often accompanied by additional diagrams, illustrations, and tables.
\item \textbf{Presentations} – presentation slides, infographics, scientific posters, memes, and all materials conveying information through text and rich visual design.
\item \textbf{Newspapers} – newspapers and magazines.
\item \textbf{Natural} – text occurring in natural environments, e.g., information boards, signs, warnings, banners, and graffiti.
\item \textbf{Business documents} – forms, contracts, regulations, financial reports, documents issued by government administration (ID cards, passports, certificates, licenses, etc.), and other types of business or official documents.
\item \textbf{Advertisements} – advertising leaflets, posters, billboards, and product packaging.
\item \textbf{Receipts \& invoices} – sales documents containing information about purchased products or services.
\item \textbf{Web \& UI} – screenshots of websites as well as desktop and mobile applications.
\end{itemize}
This categorization is also illustrated in Figure \ref{fig:document_types}, along with an example of each document type.

\begin{figure}
  \centering
  \includegraphics[scale=0.42]{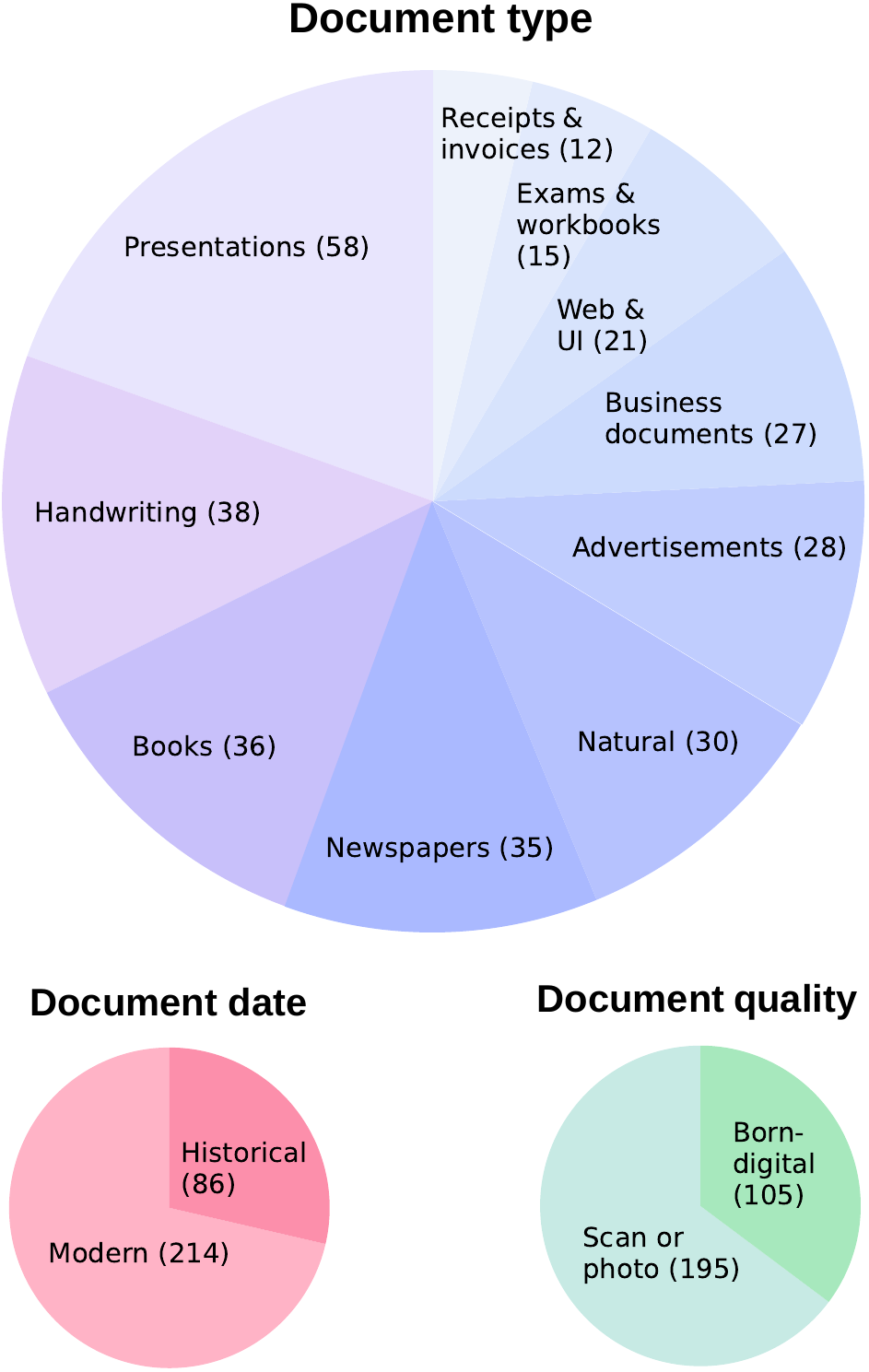}
  \caption{The number of examples in document categories (OCR, Document QA, Structured extraction), by document type, creation date, and quality.}
  \label{fig:doc_chart}
\end{figure}

Based on the time of creation, we divide the documents into \textbf{modern} (from 1990 onwards) and \textbf{historical} (up to 1989). Based on quality, we divide the documents into \textbf{born-digital} and \textbf{scans or photographs}. In Figure \ref{fig:doc_chart}, we have included three pie charts illustrating the distribution of examples within our benchmark according to each of the categorization criteria.

\subsection{Optical character recognition}

\textbf{Instruction for annotators} - This category aims to evaluate the models' ability to correctly extract text from graphic materials (such as photographs, document scans, or screenshots) and convert it into a digital format. Within this category, annotators do not create dedicated instructions for the model – they are added automatically. Annotators should focus on materials with an increased level of difficulty, which will allow testing the models' resilience to various types of distortions and visual artifacts. Particularly desirable are documents containing handwriting, including sloppy or unusual handwriting, and images suffering from such defects as: low scan resolution, insufficient lighting, blur, perspective distortions (e.g., page curvature, unusual shooting angle), as well as physical degradation of the medium (e.g., old, yellowed, or damaged paper). Furthermore, an important aspect of the evaluation is the models' ability to correctly reproduce structured data; therefore, documents containing tables should constitute a part of the test set. Verification algorithm for this category assesses the fidelity of the text transcription against the original, as well as the correctness of preserving the structure and content of tables. Each submitted sample must be provided with three tags specifying: document type, image quality, and the time of document creation.

\subsection{Document QA}

\textbf{Instruction for annotators} - This category includes questions related to documents and, more broadly, any visual objects containing text. The primary purpose of these questions is to verify the models’ ability to correctly recognize text in an image and then accurately interpret and draw conclusions from it. Therefore, when developing tasks, one should avoid requiring the model to rely on external knowledge, except in situations where it is absolutely necessary for a reliable understanding of the attached material. It is recommended to place the emphasis on constructing questions that require a multi-level understanding of the text and an analysis of the relationships between individual textual elements, as well as the accompanying graphical elements (e.g., tables, charts, infographics, or illustrations). Each submitted sample must be provided with three tags specifying: document type, image quality, and the time of document creation.

\subsection{Structured extraction}

\textbf{Instruction for annotators} - This category checks the models' ability to extract essential information from documents and output a structured JSON format compliant with a schema provided by the user. Preparing a question consists of three element: 1) Attaching a document image and an instruction for the model explaining what the document is and what specific data should be extracted from it; 2) Selecting the "Structured output" option and defining the JSON schema of the expected response; 3) Defining the ground truth JSON object. The "Structured output" panel allows specifying the fields that should be included in the resulting objects. Each field should be described by its name, data type, and a brief description regarding the field's purpose and format.
The domain of document data extraction is characterized by a high degree of diversity. When designing tasks for the benchmark, it is worth considering the following scenarios and difficulty levels:
\begin{itemize}[wide,labelwidth=0pt,labelindent=0pt,itemsep=0pt,topsep=5pt]
\item \emph{Practical applications:} Tasks may imitate commercial document digitization processes, including data extraction from receipts, invoices, financial reports, or complex accounting documentation.
\item \emph{Structured documents:} Materials that possess an explicit internal logical structure (lists, tables, sections). The model's task is then to correctly map this structure onto the resulting object.
\item \emph{Continuous text:} Tasks based on unformatted narratives that require the model to extract structured information directly from the content.
\item \emph{Documents with rich visual content:} A challenge for the model may be the extraction of data contained in various elements of a document (e.g., some fields of the resulting object are located in a table or on a chart, while others are in the text content).
\end{itemize}
Each submitted sample must be provided with three tags specifying: document type, image quality, and the time of document creation.

\section{Extended discussion of results}
\label{app:extended_results}

\textbf{Refusals of OpenAI models} - OpenAI models were the only evaluated systems for which refusals had a noticeable impact on the results. In our selective analysis of outputs from other model families, including Gemini, Grok, Qwen3.5, and Claude, refusals occurred only incidentally. We consider the affected benchmark questions safe: models should be able to answer questions about public and historical figures or works of art. We therefore treated these cases as errors caused by an overly restrictive activation of safety mechanisms.

For GPT-Audio, we identified 15 refusals in Speech QA, 10 in Sound \& Music QA, and 3 in ASR. In the image categories, GPT-5.4 produced 11 refusals in Contemporary Life, 5 in History \& Culture, and 1 in Geography \& Environment, compared with 7, 3, and 1, respectively, for GPT-5.5. No refusals occurred in the document categories. The lower refusal rate of GPT-5.5 relative to GPT-5.4 may indicate less restrictive safety behavior. Eliminating refusals would improve the OpenAI models' scores, but would not close the overall gap to Gemini-3.1-Pro, particularly in the audio categories.

\textbf{Challenges in non-speech audio understanding} - Sound \& Music QA was the most challenging category, with the best model achieving only 56\%. Our qualitative analysis indicates that most errors resulted from hallucinated interpretations of the recordings. For GPT-Audio models, refusals were also a substantial source of errors. Models from the Gemini, GPT-Audio, and MiMo families demonstrated some ability to understand non-speech audio. Voxtral, however, sometimes failed even at basic sound discrimination, for example confusing natural sounds with vehicle noise.

None of the evaluated models achieved satisfactory performance, suggesting that non-speech audio may not be sufficiently represented in their training data. This observation is consistent with results reported for MMAU, a benchmark covering speech, environmental sounds, and music, where the strongest evaluated models achieved scores of approximately 50\% \citep{sakshi2025mmau}.

\textbf{Generation temperature} - Our evaluation framework includes an automatic mechanism for adjusting the generation temperature. Requests are initially sent with a temperature of 0 whenever the model API permits it. At low temperatures, some reasoning models may enter repetitive reasoning loops, exhaust the generation budget, and return an empty final answer. When this occurs, the task is restarted with the temperature increased by 0.3. This process is repeated until a valid response is obtained. This behavior was rarely observed for frontier models or large open-weight models. In our experiments, the risk of infinite reasoning loop increased as model size decreased. It occurred regularly for smaller Qwen3.5 variants, which generate long reasoning traces by default. Smaller Gemma 4 were also affected.

\textbf{Strict and soft scoring} - We use score binarization to define task-level success: the strict score indicates whether a task was solved according to its  correctness criteria. For the OCR and ASR categories, this involved setting a threshold value above which the task would be considered solved correctly. We adopted a threshold of 0.9 as a conservative criterion. In practical ASR or OCR applications, a word error rate of 10\% is often the maximum acceptable level, and depending on the use case, the requirements may be even stricter. In our opinion, such binary values are easier to interpret when comparing across heterogeneous task types. Soft scores correspond directly to metric values before binarization and provide information about partially correct responses. However, their distributions differ across verification rules and task types. For example, OCR and Structured Extraction scores are concentrated toward 1.0, whereas scores in the QA categories are distributed more evenly. Direct comparisons of soft scores across categories should therefore be made with caution. Nevertheless, because soft scores may be more informative for some readers, we report both strict and soft results throughout the paper.

\textbf{Document categories} - The document categories classify files according to additional criteria such as date, quality, and document type. We examined the performance of selected representative models across these criteria. Our analysis included GPT-5.5, Gemini-3.1-Pro, Gemini-3-Flash, MiMo-V2.5, Gemma-4-31B, and Qwen3.5-397B-A17B. Document date had no consistent effect: three models performed better on modern documents and three on historical ones, providing no evidence that historical materials are generally more difficult. Document quality showed a clearer pattern, as all six models achieved lower overall scores on scans and photographs than on born-digital documents. This difference was particularly consistent in Structured Extraction. Results by document type should be interpreted cautiously because some groups contain few examples. Averaged across models, books achieved the highest score (78.7\%), followed by advertisements, newspapers, and business documents (approximately 73--74\%). Receipts and invoices (56.1\%) and handwriting (60.5\%) were the most challenging. Performance on handwriting also varied substantially between models, indicating that handwritten document understanding remains strongly model-dependent.

\textbf{Audio categories} - We conducted an in-depth analysis of the audio categories for six models: GPT-Audio, GPT-4o-Audio, MiMo-V2.5, Gemini-3.1-Pro, Gemini-3-Flash, and Gemini-2.5-Pro. We compared mean strict scores by the number of speakers and recording quality. The number of speakers primarily affected ASR: the average score was 72.2\% for single-speaker recordings and 47.1\% for recordings with multiple speakers. In Speech QA, however, the corresponding scores were nearly identical. This suggests that multiple speakers make precise transcription substantially more difficult, while having less impact on general understanding of the recording. Recording quality showed a consistent effect in both categories: high-quality audio improved ASR and Speech QA scores by approximately 16-17 percentage points. Sound \& Music QA did not exhibit the same pattern. Thus, recording fidelity does not appear to be the primary factor limiting non-speech audio understanding. The content and difficulty of individual tasks likely play a greater role.

\section{Error analysis}
\label{app:error_analysis}

Deterministic verification rules may produce false negatives when a semantically correct answer does not match the predefined validation criteria. To estimate the frequency of such cases, we conducted an additional manual audit of evaluation logs from four representative runs: Gemini-3.1-Pro, Gemini-3-Flash, MiMo-V2.5, and the GPT-5.5 + GPT-Audio model pair. The analysis covered the six question-answering categories: History \& Culture, Contemporary Life, Geography \& Environment, Document QA, Speech QA, and Sound \& Music QA. Each run therefore included 600 questions. We manually inspected every response with a strict score of 0 and assigned each failure to one of three categories:

\begin{itemize}[wide,labelwidth=0pt,labelindent=0pt,itemsep=0pt,topsep=5pt,parsep=0pt]
    \item \textbf{Model} - genuine model errors, including incorrect, incomplete, hallucinated, or off-topic answers;
    \item \textbf{Refusal} - cases in which the model refused to answer the question;
    \item \textbf{Rules} - semantically correct answers rejected because the verification rules were incomplete or overly restrictive.
\end{itemize}

\begin{table}[ht]
    \centering
    \small
    \setlength{\tabcolsep}{4pt}
    \begin{tabular}{l|rrr}
        \toprule
        \textbf{Model name} & \textbf{Model} & \textbf{Refusal} & \makebox[1.8cm]{\textbf{Rules}} \\
        \midrule
        Gemini-3.1-Pro        & 115 & 2  & 3 (0.5\%) \\
        GPT-5.5 + GPT-Audio   & 181 & 36 & 6 (1.0\%) \\
        Gemini-3-Flash        & 211 & 4  & 4 (0.7\%) \\
        MiMo-V2.5             & 421 & 0  & 4 (0.7\%) \\
        \bottomrule
    \end{tabular}
    \caption{Manual classification of evaluation errors. The Rules rate is calculated relative to all 600 QA tasks in each run.}
    \label{tab:error_analysis}
\end{table}

For Gemini-3.1-Pro, rule-related false negatives comprised two cases in History \& Culture and one in Geography \& Environment. For the GPT model pair, they included three cases in Geography \& Environment, two in History \& Culture, and one in Contemporary Life. Gemini-3-Flash produced one such case in each of Contemporary Life, Document QA, Geography \& Environment, and Sound \& Music QA. For MiMo-V2.5, two cases occurred in Sound \& Music QA, one in History \& Culture, and one in Speech QA.

Across individual runs, rule-related false negatives affected between 3 and 6 out of 600 questions, corresponding to approximately 0.5-1.0\% of all QA tasks. In total, we identified 17 such cases. This confirms that deterministic verification can occasionally reject valid answers, but suggests that its impact on the reported scores is limited. Most failed responses were genuine model errors. The GPT model pair also produced substantially more refusals than the other evaluated systems, which is consistent with the observation already discussed in the paper. A viable alternative to rule-based evaluation for our benchmark could be an LLM-as-a-judge approach. However, it introduces its own challenges, including errors and biases in the judge model, sensitivity to prompts and model versions, and increased evaluation costs \citep{panickssery2024llm, shi2025judging, tan2025judgebench}.

\section{Evaluation on visual categories}
\label{app:vlm_evaluation}

\begin{table*}[!t]
  \centering
  \small
  \setlength{\tabcolsep}{1pt}
  \renewcommand{\arraystretch}{1.04}
  \begin{tabular}{l|l|>{\centering\arraybackslash}m{1.1cm}|>{\centering\arraybackslash}m{1.1cm}|>{\centering\arraybackslash}m{1.1cm}|>{\centering\arraybackslash}m{1.1cm}|>{\centering\arraybackslash}m{1.1cm}|>{\centering\arraybackslash}m{1.1cm}!{\vrule width 1.06pt}>{\centering\arraybackslash}m{1.4cm}}
  \toprule
     & & \multicolumn{3}{c|}{\textbf{Images}} & \multicolumn{3}{c!{\vrule width 1.06pt}}{\textbf{Documents}} & \\
    \Xcline{3-8}{0.5pt}
    \multirow{-2}{*}{\textbf{Model}} & \multirow{-2}{*}{\textbf{Reference}} & {\scriptsize History \& culture} & {\scriptsize Contemp. life} & {\scriptsize Geography \& env.} & {\scriptsize OCR} & {\scriptsize Document QA} & {\scriptsize Structured extraction}  & \multirow{-2}{*}{\makecell[c]{\textbf{Mean} \\ \textbf{score}}} \\[-1pt]
    \midrule
Gemini-2.5-Flash & \citet{comanici2025gemini} & 50 \softscore{58.2} & 40 \softscore{51.0} & 53 \softscore{66.2} & 73 \softscore{91.0} & 52 \softscore{64.4} & 53 \softscore{91.9} & 53.5 \softscore{70.4} \\
Gemini-3.1-Flash-Lite & \citet{google2025gemini3} & 58 \softscore{63.2} & 49 \softscore{58.2} & 62 \softscore{76.2} & 76 \softscore{91.1} & 44 \softscore{60.0} & 45 \softscore{91.0} & 55.7 \softscore{73.3} \\
Gemini-3-Flash & \citet{google2025gemini3} & 70 \softscore{76.2} & 74 \softscore{80.2} & 69 \softscore{79.7} & 84 \softscore{92.5} & 64 \softscore{76.8} & 68 \softscore{96.7} & 71.5 \softscore{83.7} \\
Gemini-2.5-Pro & \citet{comanici2025gemini} & 70 \softscore{76.0} & 72 \softscore{80.2} & 72 \softscore{80.8} & 80 \softscore{92.3} & 72 \softscore{83.6} & 68 \softscore{96.4} & 72.3 \softscore{84.9} \\
Gemini-3.5-Flash & \citet{google2025gemini3} & 82 \softscore{85.8} & 79 \softscore{85.3} & \globalbest{\groupbest{86} \softscore{92.6}} & \groupbest{86} \softscore{94.3} & \globalbest{\groupbest{88} \softscore{94.7}} & 66 \softscore{94.9} & 81.2 \softscore{91.3} \\
Gemini-3.1-Pro & \citet{google2025gemini3} & \globalbest{\groupbest{82} \softscore{86.9}} & \globalbest{\groupbest{83} \softscore{87.3}} & 85 \softscore{91.1} & 81 \softscore{91.4} & 83 \softscore{89.7} & \globalbest{\groupbest{79} \softscore{97.3}} & \globalbest{\groupbest{82.2} \softscore{90.6}} \\[-1pt]
\midrule
Claude-Haiku-4.5 & \citet{anthropic2025claudehaiku45} & 20 \softscore{28.5} & 9 \softscore{15.4} & 21 \softscore{30.8} & 22 \softscore{61.5} & 11 \softscore{29.2} & 19 \softscore{73.4} & 17.0 \softscore{39.8} \\
Claude-Sonnet-4.6 & \citet{anthropic2026claudesonnet46} & 54 \softscore{58.8} & 26 \softscore{34.9} & 53 \softscore{64.0} & 79 \softscore{92.5} & 49 \softscore{66.0} & 52 \softscore{92.8} & 52.2 \softscore{68.2} \\
Claude-Opus-4.7 & \citet{anthropic2026claudeopus47} & 62 \softscore{69.9} & 36 \softscore{44.4} & 62 \softscore{74.0} & 88 \softscore{94.9} & 63 \softscore{73.9} & 59 \softscore{94.6} & 61.7 \softscore{75.3} \\
Claude-Fable-5 & \citet{anthropic2026claude} & \groupbest{76} \softscore{79.0} & \groupbest{62} \softscore{70.1} & \groupbest{85} \softscore{89.8} & \globalbest{\groupbest{90} \softscore{94.2}} & \groupbest{85} \softscore{91.2} & \groupbest{70} \softscore{96.0} & \groupbest{78.0} \softscore{86.7} \\ [-1pt]
\midrule
GPT-5.4-Nano & \citet{singh2025openai} & 20 \softscore{27.6} & 20 \softscore{25.6} & 22 \softscore{33.0} & 17 \softscore{43.7} & 28 \softscore{43.8} & 11 \softscore{59.5} & 19.7 \softscore{38.9} \\
GPT-4o & \citet{hurst2024gpt} & 41 \softscore{50.7} & 39 \softscore{47.2} & 48 \softscore{58.9} & 17 \softscore{31.8} & 23 \softscore{44.4} & 25 \softscore{80.2} & 32.2 \softscore{52.2} \\
GPT-5.4-Mini & \citet{singh2025openai} & 47 \softscore{52.1} & 35 \softscore{40.5} & 50 \softscore{61.2} & 60 \softscore{84.7} & 62 \softscore{72.7} & 36 \softscore{85.5} & 48.3 \softscore{66.1} \\
GPT-5.4 & \citet{singh2025openai} & 47 \softscore{56.1} & 42 \softscore{52.8} & 55 \softscore{64.3} & 71 \softscore{89.7} & 54 \softscore{67.2} & 45 \softscore{91.4} & 52.3 \softscore{70.3} \\
GPT-5.5 & \citet{singh2025openai} & \groupbest{74} \softscore{79.2} & \groupbest{73} \softscore{79.0} & \groupbest{77} \softscore{84.5} & \groupbest{77} \softscore{92.3} & \groupbest{83} \softscore{91.3} & \groupbest{55} \softscore{94.5} & \groupbest{73.2} \softscore{86.8} \\[-1pt]
\midrule
Kimi-K2.6 & \citet{team2026kimi} & 50 \softscore{59.9} & 40 \softscore{52.6} & 47 \softscore{60.6} & 67 \softscore{89.6} & 71 \softscore{82.7} & 46 \softscore{91.2} & 53.5 \softscore{72.8} \\
Kimi-K2.5 & \citet{team2026kimi} & 58 \softscore{65.3} & 44 \softscore{56.5} & 52 \softscore{64.8} & 73 \softscore{90.1} & 65 \softscore{80.0} & 45 \softscore{92.3} & 56.2 \softscore{74.8} \\
Kimi-K3 & \citet{moonshotai2026k3} & \groupbest{55} \softscore{61.8} & \groupbest{57} \softscore{64.9} & \groupbest{60} \softscore{71.8} & \groupbest{80} \softscore{92.6} & \groupbest{79} \softscore{89.1} & \groupbest{54} \softscore{93.8} & \groupbest{64.2} \softscore{79.0} \\[-1pt]
\midrule
Qwen3.5-2B & \citet{qwen3.5} & 2 \softscore{6.5} & 4 \softscore{9.5} & 2 \softscore{10.9} & 28 \softscore{69.2} & 5 \softscore{18.5} & 12 \softscore{65.6} & 8.8 \softscore{30.0} \\
Qwen3.5-4B & \citet{qwen3.5} & 11 \softscore{16.9} & 6 \softscore{12.8} & 7 \softscore{17.0} & 42 \softscore{78.4} & 39 \softscore{54.3} & 23 \softscore{82.3} & 21.3 \softscore{43.6} \\
Qwen3.5-9B & \citet{qwen3.5} & 10 \softscore{15.0} & 10 \softscore{16.9} & 8 \softscore{17.1} & 55 \softscore{76.9} & 50 \softscore{59.4} & 30 \softscore{66.4} & 27.2 \softscore{41.9} \\
Qwen3.6-27B & \citet{qwen3.6-27b} & 22 \softscore{31.1} & 19 \softscore{25.7} & 19 \softscore{32.7} & 61 \softscore{81.5} & 62 \softscore{78.0} & 36 \softscore{70.7} & 36.5 \softscore{53.3} \\
Qwen3.5-27B & \citet{qwen3.5} & 24 \softscore{31.1} & 16 \softscore{19.1} & 18 \softscore{30.1} & 66 \softscore{85.9} & 63 \softscore{72.5} & \groupbest{37} \softscore{89.0} & 37.3 \softscore{54.6} \\
Qwen3.5-35B-A3B & \citet{qwen3.5} & 23 \softscore{31.3} & 20 \softscore{26.1} & 22 \softscore{33.0} & 67 \softscore{85.9} & 61 \softscore{72.8} & 33 \softscore{68.0} & 37.7 \softscore{52.9} \\
Qwen3.6-35B-A3B & \citet{qwen36_35b_a3b} & 23 \softscore{31.1} & 21 \softscore{27.8} & 22 \softscore{35.9} & 69 \softscore{89.3} & 60 \softscore{72.3} & 36 \softscore{89.2} & 38.5 \softscore{57.6} \\
Qwen3.5-122B-A10B & \citet{qwen3.5} & 35 \softscore{44.3} & 28 \softscore{36.4} & 29 \softscore{44.2} & 73 \softscore{90.2} & 64 \softscore{76.0} & 36 \softscore{88.1} & 44.2 \softscore{63.2} \\
Qwen3.5-397B-A17B & \citet{qwen3.5} & \groupbest{51} \softscore{56.5} & \groupbest{40} \softscore{49.9} & \groupbest{51} \softscore{63.3} & \groupbest{78} \softscore{92.5} & \groupbest{78} \softscore{87.8} & 35 \softscore{89.5} & \groupbest{55.5} \softscore{73.2} \\[-1pt]
\midrule
Gemma-4-E2B & \citet{farabet2026gemma4} & 4 \softscore{6.7} & 4 \softscore{9.4} & 4 \softscore{14.1} & 30 \softscore{72.3} & 21 \softscore{34.2} & 15 \softscore{73.2} & 13.0 \softscore{35.0} \\
Gemma-4-E4B & \citet{farabet2026gemma4} & 9 \softscore{15.8} & 10 \softscore{14.6} & 8 \softscore{17.6} & 42 \softscore{80.4} & 35 \softscore{51.3} & 21 \softscore{77.8} & 20.8 \softscore{42.9} \\
Gemma-4-26B-A4B & \citet{farabet2026gemma4} & \groupbest{39} \softscore{48.1} & 26 \softscore{35.4} & 32 \softscore{46.6} & 62 \softscore{87.3} & 57 \softscore{72.2} & 40 \softscore{89.7} & 42.7 \softscore{63.2} \\
Gemma-4-31B & \citet{farabet2026gemma4} & 36 \softscore{45.7} & \groupbest{32} \softscore{42.9} & \groupbest{47} \softscore{57.7} & \groupbest{76} \softscore{92.3} & \groupbest{70} \softscore{82.3} & \groupbest{59} \softscore{95.0} & \groupbest{53.3} \softscore{69.3} \\[-1pt]
\midrule
Grok-4.1-Fast & \citet{xai2025grok41modelcard} & 36 \softscore{45.6} & 32 \softscore{43.3} & 39 \softscore{50.7} & 1 \softscore{30.6} & 28 \softscore{44.4} & 6 \softscore{54.8} & 23.7 \softscore{44.9} \\
Grok-4.20 & \citet{xai2026grok420} & 36 \softscore{45.3} & 25 \softscore{34.0} & 39 \softscore{50.1} & 21 \softscore{32.0} & 19 \softscore{39.6} & 34 \softscore{87.4} & 29.0 \softscore{48.1} \\
Grok-4 & \citet{xai2025grok4modelcard} & 53 \softscore{62.4} & \groupbest{47} \softscore{57.2} & \groupbest{55} \softscore{67.8} & 17 \softscore{59.5} & 47 \softscore{64.3} & 19 \softscore{76.9} & 39.7 \softscore{64.7} \\
Grok-4.3 & \citet{xai2025grok4modelcard} & \groupbest{55} \softscore{62.4} & 42 \softscore{50.4} & 52 \softscore{64.7} & \groupbest{66} \softscore{85.9} & \groupbest{57} \softscore{71.0} & \groupbest{39} \softscore{88.7} & \groupbest{51.8} \softscore{70.5} \\[-1pt]
\midrule
MiMo-V2-Omni & \citet{xiaomi2026mimov2omni} & 21 \softscore{30.6} & 27 \softscore{33.5} & 28 \softscore{42.9} & \groupbest{57} \softscore{82.2} & 53 \softscore{68.1} & 16 \softscore{39.5} & 33.7 \softscore{49.5} \\
MiMo-V2.5 & \citet{xiaomi2026mimov25} & \groupbest{22} \softscore{30.6} & \groupbest{23} \softscore{32.5} & \groupbest{31} \softscore{42.0} & \groupbest{57} \softscore{82.8} & \groupbest{62} \softscore{72.1} & \groupbest{39} \softscore{88.7} & \groupbest{39.0} \softscore{58.1} \\[-1pt]
\midrule
GLM-4.6V & \citet{hong2025glm} & 31 \softscore{38.2} & 20 \softscore{30.2} & 23 \softscore{36.5} & \groupbest{36} \softscore{73.8} & 41 \softscore{57.7} & 12 \softscore{51.7} & 27.2 \softscore{48.0} \\
GLM-5V-Turbo & \citet{zeng2026glm} & \groupbest{45} \softscore{54.0} & \groupbest{35} \softscore{47.3} & \groupbest{42} \softscore{55.8} & 34 \softscore{73.5} & \groupbest{48} \softscore{64.2} & \groupbest{23} \softscore{81.3} & \groupbest{37.8} \softscore{62.7} \\[-1pt]
\midrule
Mistral-3.2-24B & \citet{mistralai2025mistralsmall31} & 15 \softscore{21.9} & 17 \softscore{24.9} & 7 \softscore{19.9} & 1 \softscore{1.1} & 18 \softscore{36.5} & 18 \softscore{78.8} & 12.7 \softscore{30.5} \\
Mistral-Small-4 & \citet{mistralai2026mistralsmall4} & 13 \softscore{19.9} & 14 \softscore{18.7} & 13 \softscore{21.3} & 13 \softscore{28.4} & 21 \softscore{38.5} & 18 \softscore{70.9} & 15.3 \softscore{33.0} \\
Mistral-Large-3-2512 & \citet{mistralai2025mistral3} & 24 \softscore{32.1} & 18 \softscore{24.4} & 22 \softscore{34.8} & 5 \softscore{7.9} & 19 \softscore{38.1} & 15 \softscore{78.0} & 17.2 \softscore{35.9} \\
Mistral-Medium-3.1 & \citet{mistralai2025mistralmedium3} & \groupbest{36} \softscore{44.4} & 20 \softscore{29.4} & 27 \softscore{40.4} & 6 \softscore{9.8} & 14 \softscore{34.8} & 16 \softscore{79.0} & 19.8 \softscore{39.6} \\
Mistral-Medium-3.5 & \citet{mistralai2026remoteagents} & 33 \softscore{40.2} & \groupbest{29} \softscore{38.7} & \groupbest{31} \softscore{42.2} & \groupbest{40} \softscore{74.8} & \groupbest{36} \softscore{57.2} & \groupbest{27} \softscore{80.0} & \groupbest{32.7} \softscore{55.5} \\[-1pt]
\midrule
Llama-4-Scout & \citet{adcock2026llama} & 13 \softscore{19.6} & 13 \softscore{19.9} & 17 \softscore{30.7} & 23 \softscore{65.0} & \groupbest{21} \softscore{38.7} & 18 \softscore{76.4} & 17.5 \softscore{41.7} \\
Llama-4-Maverick & \citet{adcock2026llama} & \groupbest{21} \softscore{29.8} & \groupbest{19} \softscore{28.4} & \groupbest{26} \softscore{38.5} & \groupbest{28} \softscore{68.1} & 19 \softscore{38.3} & \groupbest{19} \softscore{78.9} & \groupbest{22.0} \softscore{47.0} \\[-1pt]
\midrule
InternVL3.5-8B & \citet{wang2025internvl3_5} & 2 \softscore{6.7} & 6 \softscore{11.7} & 4 \softscore{10.4} & 1 \softscore{18.9} & 9 \softscore{22.5} & 9 \softscore{58.6} & 5.2 \softscore{21.5} \\
InternVL3.5-30B-A3B & \citet{wang2025internvl3_5} & 3 \softscore{6.4} & \groupbest{8} \softscore{12.8} & 4 \softscore{11.7} & 2 \softscore{36.2} & 8 \softscore{22.8} & 8 \softscore{61.9} & 5.5 \softscore{25.3} \\
InternVL3.5-14B & \citet{wang2025internvl3_5} & 1 \softscore{7.1} & 7 \softscore{13.8} & 5 \softscore{13.1} & 4 \softscore{30.3} & 17 \softscore{30.2} & 5 \softscore{63.4} & 6.5 \softscore{26.3} \\
InternVL3.5-38B & \citet{wang2025internvl3_5} & \groupbest{5} \softscore{10.0} & 7 \softscore{14.0} & \groupbest{11} \softscore{20.1} & \groupbest{5} \softscore{22.8} & \groupbest{20} \softscore{37.5} & \groupbest{12} \softscore{69.9} & \groupbest{10.0} \softscore{29.0} \\[-1pt]
\midrule
LLaVA-Bielik-11b-v2.6 & \citet{statkiewicz2026annotation} & 11 \softscore{19.7} & 15 \softscore{22.7} & 11 \softscore{20.9} & 2 \softscore{29.7} & 3 \softscore{13.4} & 0 \softscore{25.0} & 7.0 \softscore{21.9} \\
LLaVA-PLLuM-12b-nc & \citet{statkiewicz2026annotation} & \groupbest{18} \softscore{24.5} & \groupbest{18} \softscore{24.5} & \groupbest{12} \softscore{18.3} & \groupbest{3} \softscore{25.4} & \groupbest{5} \softscore{14.4} & \groupbest{2} \softscore{24.2} & \groupbest{9.7} \softscore{21.9} \\[-1pt]
\midrule
Phi-4-Multimodal-Instruct & \citet{abouelenin2025phi} & \groupbest{2} \softscore{6.2} & \groupbest{2} \softscore{3.4} & \groupbest{5} \softscore{11.8} & \groupbest{0} \softscore{0.6} & \groupbest{6} \softscore{11.8} & \groupbest{0} \softscore{14.8} & \groupbest{2.5} \softscore{8.1} \\[-1pt]
\bottomrule
  \end{tabular}
  \caption{Results of the evaluation of VLMs and multimodal models on visual categories. We report strict scores and soft scores (the smaller gray numbers in parentheses). The best score in each column is highlighted with light gray shading and boldface, while the highest score within each model group is underlined.}
  \label{tab:vllms}
\end{table*}

Currently, few models simultaneously support text, audio, and vision. This limits the scope of evaluation on the full benchmark. However, the number of vision language models (VLMs) supporting the Polish language is significantly larger. Therefore, we decided to conduct the evaluation solely on six vision-based categories to assess the capabilities of available models in this domain. Table \ref{tab:vllms} presents the results of over 50 models, broken down by model family.

The strongest models are found among the Gemini, GPT, and Claude families, which is consistent with the previously discussed results on the full benchmark. Following them are the Kimi \citep{team2026kimi}, Qwen3.5 / 3.6 \citep{qwen3.5,qwen3.6-27b,qwen36_35b_a3b}, and Gemma 4 \citep{farabet2026gemma4} models. Kimi models offer good image understanding capabilities, but they are very large, with over 1 trillion parameters for Kimi-K2.5/K2.6 and 2.8 trillion for Kimi-K3. The subsequent positions, however, are particularly interesting, as both the Gemma 4 and Qwen3.5 families include smaller versions of around 30B parameters that perform quite well, achieving notably high scores in document-based categories. Therefore, these models should be a good choice for document processing in local deployments. The commercial Grok models performed below expectations in image processing, especially considering that they handle the Polish language well in text-based capabilities \citep{dadas2025evaluating}. The next positions are occupied by the multimodal MiMo \citep{xiaomi2026mimov2omni,xiaomi2026mimov25} models and GLMs \citep{zeng2026glm,hong2025glm}. Next, we have the Mistral models, most of which struggle with document processing, but the latest Mistral-Medium-3.5 shows a noticeable improvement in this area. Following them are the Llama 4 models \citep{adcock2026llama}. The InternVL-3.5 \citep{wang2025internvl3_5} models perform noticeably worse than the Mistrals and Llama 4, offering poor support for the Polish language. Next, we have the Polish versions of LLaVa \citep{statkiewicz2026annotation}, which represent the first, or among the first, attempts to train VLMs for the Polish language. In their case, there is a distinct lack of capability in document processing, alongside slightly better results in knowledge-based categories. Phi-4-Multimodal-Instruct \citep{abouelenin2025phi} ranked last among the tested models.

\section{Annotation process}
\label{app:annotation_process}
The benchmark was created by a group of six annotators (four male and two female), all in their thirties or forties. All annotators are native Polish speakers who have lived in Poland since birth, ensuring strong familiarity with local language use, cultural references, and social context. The annotation process spanned from mid-October 2025 to the end of March 2026 and resulted in an average of around 40 new tasks per week across the team, accounting for annotator availability and holidays. The contribution of each annotator was approximately balanced.

Annotators followed a unified set of guidelines specifying how to construct tasks and evaluation criteria. Questions were required to be unambiguous and compatible with rule-based verification. To ensure robustness, annotators were encouraged to anticipate multiple valid answer formulations and encode them within evaluation conditions (e.g., through alternative acceptable expressions or flexible matching rules). The guidelines also emphasized enforcing concise answers, combining inclusion and exclusion criteria for closed questions, and systematically testing tasks against model-generated responses during creation.

Task creation was performed with explicit consideration of cultural grounding. Annotators incorporated references to Polish history, public figures, traditions, language use, and everyday life, often requiring the model to connect multimodal input with culturally specific knowledge.

The annotation process also included deliberate control of task difficulty. Annotators constructed both straightforward recognition-based questions and more complex tasks requiring reasoning, information integration, or contextual interpretation. This balance was achieved iteratively, with ongoing refinement based on observed model behavior.

Additionally, annotators monitored the distribution of tasks across tags. As the dataset grew, they introduced questions from less represented areas to avoid strongly underrepresented subcategories and ensure sufficient diversity.

\subsection{Review and Revision Phase}

Once the initial annotation process was completed, the dataset underwent a dedicated review and revision phase. This stage involved three annotators, all native Polish speakers, who were responsible for identifying issues in task design, content quality, and evaluation criteria.

To support this process, we conducted large-scale evaluations using multiple multimodal models, including Gemini-3.1-Pro, Gemini-3-Flash, GPT-5.4, and GPT-Audio. Annotators analyzed evaluation logs to detect inconsistencies, ambiguities, and potential failure cases in both question formulation and verification rules. Identified issues were subsequently discussed and resolved through iterative refinement. The purpose of the review phase was not to calibrate question difficulty. No questions were modified or removed based on whether the models were able to answer them. The annotators responsible for the review followed predefined task acceptance criteria. These criteria were as follows:
\begin{itemize}[wide,labelwidth=0pt,labelindent=0pt,itemsep=0pt,topsep=5pt,parsep=0pt]
\item A question should be formulated in such a way that unambiguous verification rules can be prepared for it. If a question was too general or ambiguous, it was removed or modified to better fit our rule-based framework.
\item All questions in the image categories, as well as in the Sound \& Music QA category, should be grounded in Polish culture. In the remaining categories, questions requiring only language or speech understanding were allowed.
\item Questions should require analysis of the attached multimodal input. If a question could be answered solely on the basis of the textual prompt, it was removed or modified.
\item A question should not duplicate another question. Questions that differed in wording but used the same multimodal input were also treated as duplicates.
\item A question should not concern ephemeral knowledge that quickly becomes outdated or cultural phenomena whose lifetime is shorter than a few years.
\end{itemize}

As a result of this phase, 28 tasks were removed and replaced with newly created ones, while 57 tasks underwent substantial revisions, such as rewriting the question prompt or modifying the evaluation conditions. Minor editorial corrections, including stylistic improvements and typo fixes, were applied continuously throughout the process.

Overall, this phase played a critical role in improving dataset consistency, ensuring stronger alignment with the benchmark’s objectives, and enhancing the reliability of automatic evaluation.

\begin{figure}
  \centering
  \includegraphics[scale=0.40]{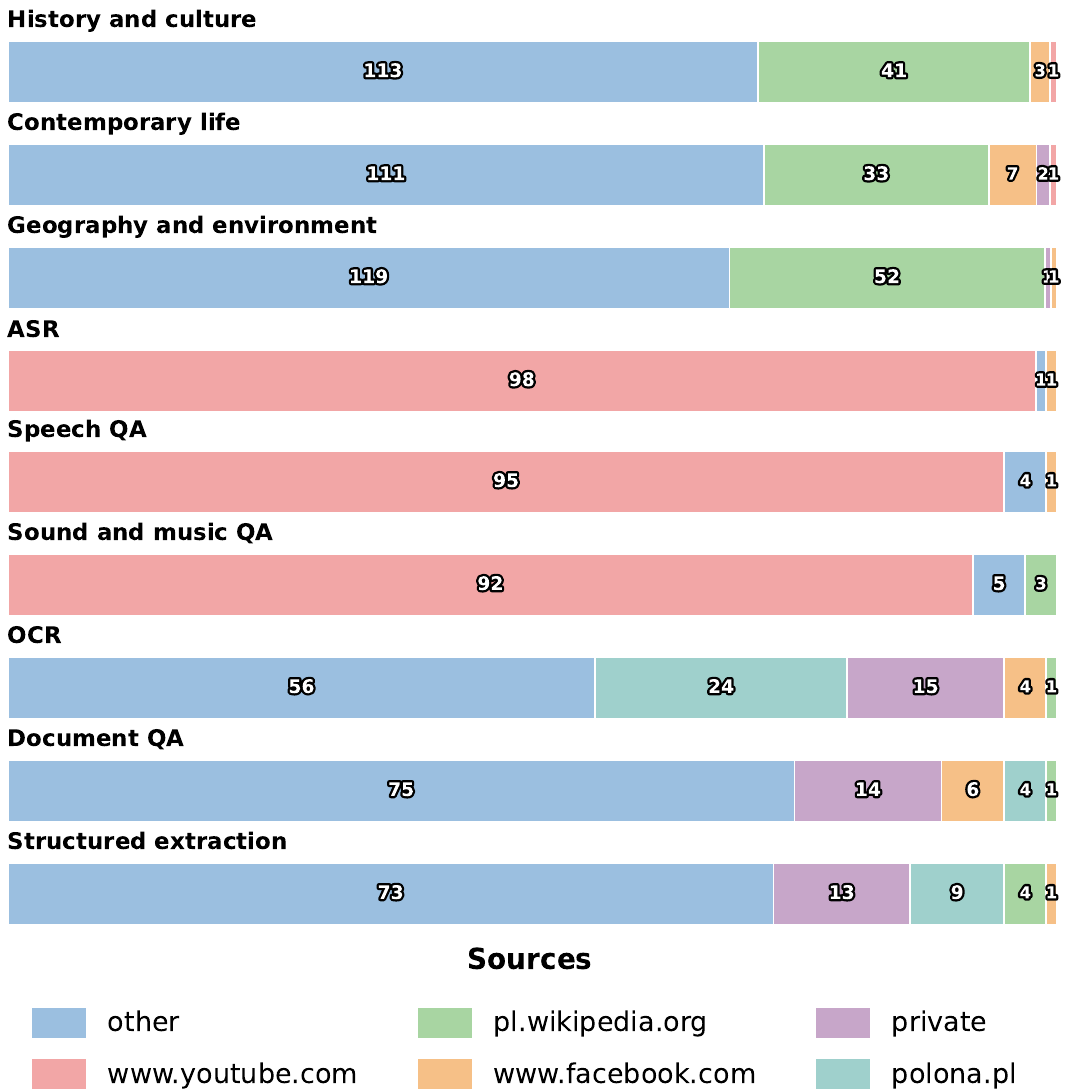}
  \caption{A chart showing the distribution of files according to the most common sources we used. The \emph{private} source refers to files created by the annotators themselves.}
  \label{fig:data_sources}
\end{figure}

\subsection{Data Sources}
\label{app:data_sources}
Our annotation framework allows annotators to attach provenance metadata to every source file used in a task. These metadata include the original source URL, the name of the original author or creator, and the license under which the file was made available. Figure~\ref{fig:data_sources} presents the distribution of source files across benchmark categories and the most frequently represented source domains. The \emph{other} group aggregates less frequent domains, while \emph{private} denotes files created by the annotators. Counts do not necessarily sum to 100 within each category because some files may have more than one source. Examples of questions with multiple sources can be found in Appendix \ref{app:examples} (questions 3, 4, and 5).

\section{Annotation app}
\label{app:application}
The process of constructing a benchmark from scratch is inherently complex and time-consuming, requiring effective coordination of annotators and careful control of task quality. To support this process, we developed a dedicated application for creating, managing, reviewing, and evaluating multimodal tasks. The application records all edits made by annotators. It currently supports nine languages: Polish, English, German, French, Dutch, Portuguese, Spanish, Italian, and Russian. Support for additional languages can be added easily, as the only language-dependent components are the lemmatizer and the default prompts used for the ASR, OCR, and Structured Extraction categories.

The application is organized into several modules, responsible, among others, for model configuration, annotator statistics, question management, and the question editing interface.

\textbf{The model management module} allows to add and modify models used in the evaluation process. Each model is defined by a set of parameters in a structured JSON format, allowing for flexible configuration of properties such as endpoints, maximum number of tokens, and required prefixes. The interface presents a list of available models with their identifiers and supports editing and deleting operations. A key feature of this module is the ability to assign models to specific question categories.

The application also offers \textbf{a statistics module} for monitoring annotator activity and tracking overall progress in building the dataset. The interface includes a weekly summary of completed tasks, allowing for a clear comparison of individual contributions over time. This facilitates ongoing monitoring of the annotation process and helps identify potential bottlenecks or workload imbalances. Furthermore, a task overview panel allows users to track the number of tasks created during the current week, providing up-to-date insight into the dataset's development. Annotators can filter and browse tasks by category, tag, and each user has access to a personalized view of their tasks. This supports self-monitoring of progress and efficient workflow.

\textbf{The task overview panel} allows you to view and manage all tasks in the dataset. It includes category-based filters, each tagged with the number of tasks assigned to that category, allowing for quick navigation and data segmentation. Additional controls allow users to switch to personalized views, such as tasks assigned to the current user or tasks created this week, as well as filter untagged items. Tasks can also be searched for by category name or content, streamlining searches for specific entries. The main section consists of a table listing tasks, where each row represents a single question. The table includes information such as the category, input content, assigned tags, the annotator responsible for the task, and the creation date. Action controls are also available for each entry, allowing for quick task management, such as deleting.

\begin{figure*}[!htbp]
  \centering
  \includegraphics[scale=0.55]{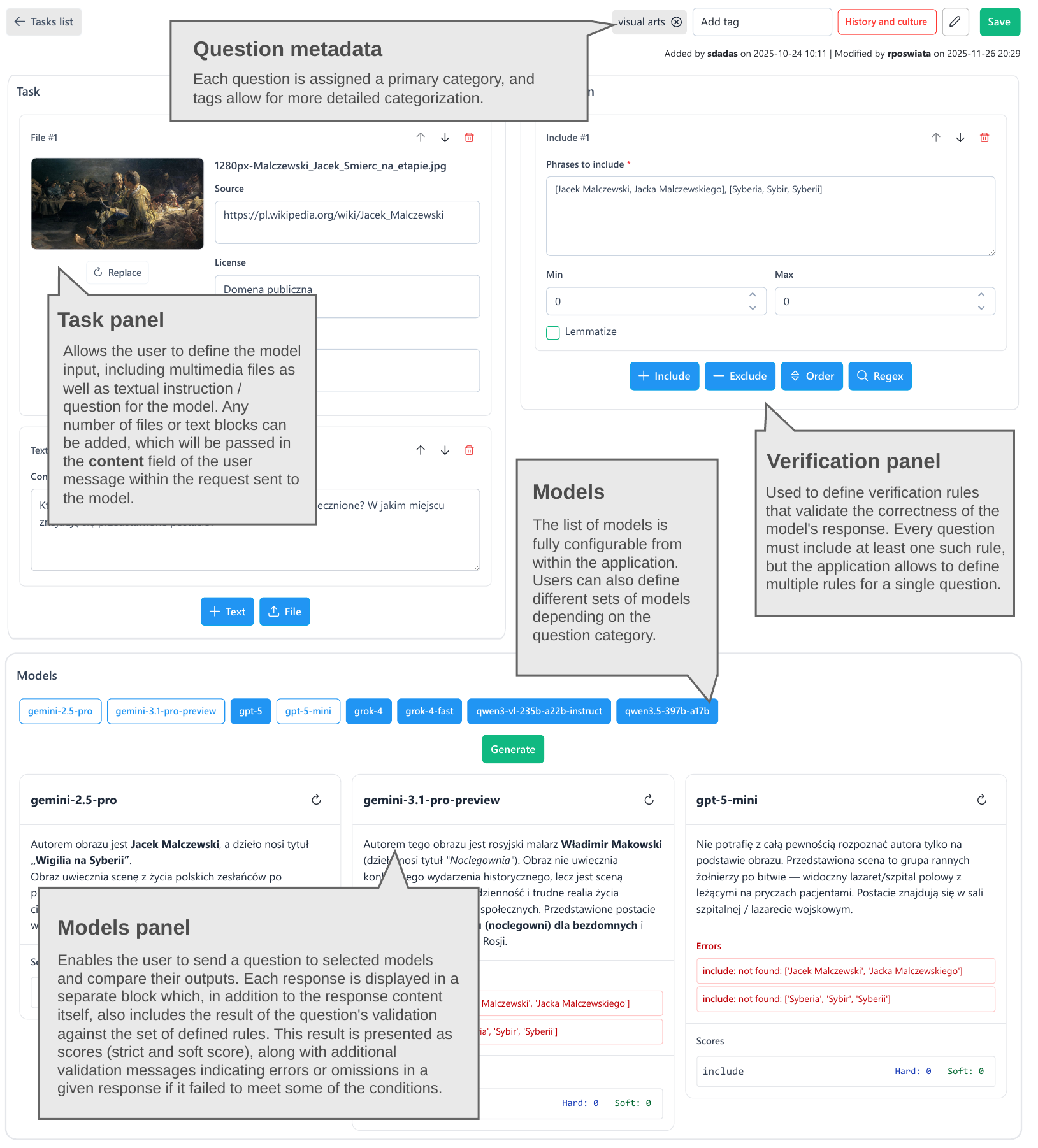}
  \caption{A sample screenshot of the task edit view in our annotator app, along with a description of the available features.}
  \label{fig:app}
\end{figure*}

\textbf{The view for adding and editing tasks} is a key feature of the app. This is where the task is defined, which will later be used to evaluate models. This screen is shown in Figure \ref{fig:app}, along with a description of its most important features. Users can define model inputs, such as images, audio, or text, and add metadata such as license information, author, and source. They can also formulate any question or command, depending on the category to which the question will be assigned. An important element is the verification section, where the conditions that a correct answer should meet are defined (e.g. required phrases), which allows for automatic evaluation of the results. This screen offers considerable flexibility, both in terms of the task content and the selection of models for which the answer will be evaluated. Model responses are placed in separate blocks, one for each selected model. This block contains information including the model name, the received response, and the result of validation rules. This allows for ongoing tracking of model responses and oversight of the dataset building process.

\clearpage
\onecolumn
\section{Sample questions}
\label{app:examples}

{
\small
\begin{longtable}{|>{\centering\arraybackslash}m{4cm}m{11cm}|}
\hline
\vspace{5px}\includegraphics[width=4cm]{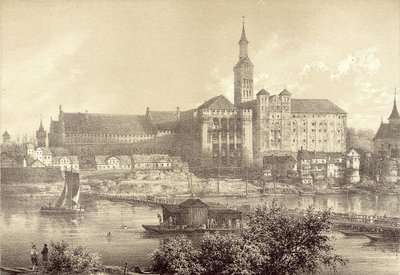} & \vspace{5px}\textbf{Category:} History and culture\vspace{5px}\newline\textbf{Instruction:} Na litografii ukazano jedną z największych średniowiecznych warowni Europy. Na podstawie przedstawienia rozpoznaj ją i podaj jej budowniczych oraz rok przejęcia przez Królestwo Polskie.\vspace{5px}\newline\textbf{Instruction (english translation):} The lithograph depicts one of the largest medieval fortresses in Europe. Based on the depiction, identify it and provide its builders, as well as the year it was taken over by the Kingdom of Poland.\vspace{5px}\newline\textbf{Verification rules:} \textbf{Include}([Krzyżacy, Zakon Krzyżacki], 1457)\vspace{5px}\\
\hline
\vspace{5px}\includegraphics[width=4cm]{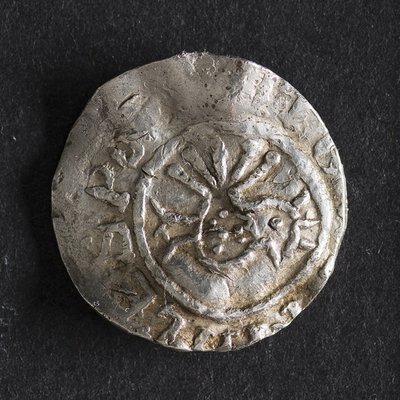} & \vspace{5px}\textbf{Category:} History and culture\vspace{5px}\newline\textbf{Instruction:} W mennicy którego polskiego władcy została wybita moneta widoczna na fotografii?\vspace{5px}\newline\textbf{Instruction (english translation):} In which Polish ruler's mint was the coin shown in the photograph struck?\vspace{5px}\newline\textbf{Verification rules:} \textbf{Include}([Bolesław Chrobry, Bolesław I Chrobry])\vspace{5px}\\
\hline
\vspace{5px}\includegraphics[width=4cm]{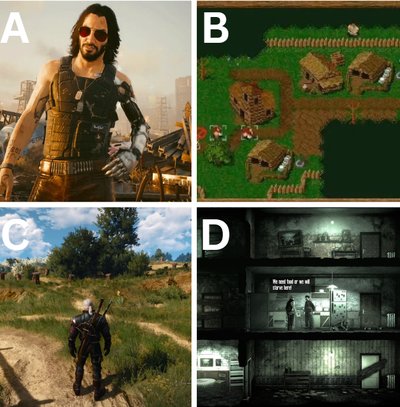} & \vspace{5px}\textbf{Category:} Contemporary life\vspace{5px}\newline\textbf{Instruction:} Obrazy oznaczone literami A, B, C, D przedstawiają kadry z polskich gier komputerowych. Czy jesteś w stanie rozpoznać te gry? Podaj tytuły w kolejności od A do D, bez dodatkowych komentarzy.\vspace{5px}\newline\textbf{Instruction (english translation):} The images marked with letters A, B, C, D show screenshots from Polish video games. Can you recognize these games? Provide the titles in order from A to D, without any additional comments.\vspace{5px}\newline\textbf{Verification rules:} \textbf{Order}(Cyberpunk, Polanie, Wiedźmin, This War of Mine)\vspace{5px}\\
\hline
\vspace{5px}\includegraphics[width=4cm]{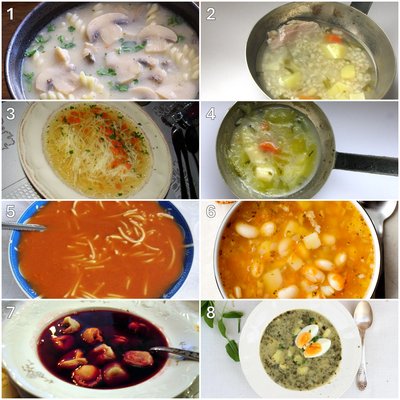} & \vspace{5px}\textbf{Category:} Contemporary life\vspace{5px}\newline\textbf{Instruction:} Podaj nazwy przedstawionych na zdjęciach zup.  Zwróć tylko listę w formacie [numer zdjęcia] - [nazwa zupy], bez dodatkowych komentarzy.\vspace{5px}\newline\textbf{Instruction (english translation):} Provide the names of the soups shown in the photos. Return only a list in the format [photo number] - [soup name], without any additional comments.\vspace{5px}\newline\textbf{Verification rules:} \textbf{Order}(pieczarkowa, krupnik, rosół, ogórkowa, pomidorowa, fasolowa, [barszcz czerwony, barszcz z uszkami], szczawiowa)\vspace{5px}\\
\hline
\vspace{5px}\includegraphics[width=4cm]{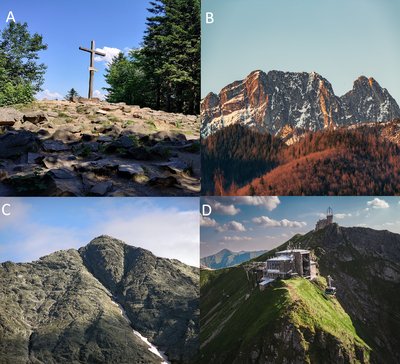} & \vspace{5px}\textbf{Category:} Geography and environment\vspace{5px}\newline\textbf{Instruction:} Wyodrębnij i uszereguj widoczne na powyższym zdjęciu, charakterystyczne polskie szczyty górskie w kolejności od najniższego do najwyższego. Każda z gór została oznaczona literą A, B, C lub D.  Odpowiedz tylko ciągiem liter A, B, C, D w odpowiedniej kolejności.\vspace{5px}\newline\textbf{Instruction (english translation):} Identify and sort the characteristic Polish mountain peaks visible in the photo above from lowest to highest. Each mountain has been marked with the letter A, B, C, or D. Answer only with a sequence of the letters A, B, C, D in the correct order.\vspace{5px}\newline\textbf{Verification rules:} \textbf{Order}(A, B, D, C)\vspace{5px}\\
\hline
\vspace{5px}\includegraphics[width=4cm]{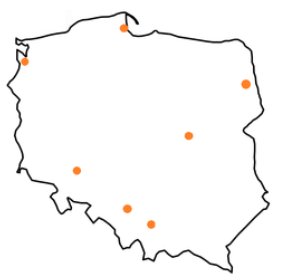} & \vspace{5px}\textbf{Category:} Geography and environment\vspace{5px}\newline\textbf{Instruction:} Podaj nazwy miast zaznaczonych na powyższej mapie. Posortuj ich nazwy według kierunku północ-południe zaczynając od miasta wysuniętego najbardziej na północ. Podaj tylko posortowną listę miast i nie uwzględniaj dodatkowego komentarza.\vspace{5px}\newline\textbf{Instruction (english translation):} Provide the names of the cities marked on the map above. Sort their names from north to south, starting with the northernmost city. Provide only the sorted list of cities and do not include any additional comments.\vspace{5px}\newline\textbf{Verification rules:} \textbf{Order}(Gdańsk, Szczecin, Białystok, Warszawa, Wrocław, Katowice, Kraków)\vspace{5px}\\
\hline
\vspace{5pt}\includegraphics[width=4cm]{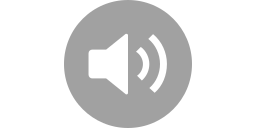}\vspace{5pt} \newline \emph{\textbf{File description: }An explanatory recording about how a moka pot works and how to use it.} & \vspace{5px}\textbf{Category:} ASR\vspace{5px}\newline\textbf{Instruction:} [General prompt for ASR]\vspace{5px}\newline\textbf{Verification rules:} \textbf{WAcc}(
\begin{adjustwidth}{0.5cm}{0cm}
Jak zaparzyć kawę w kawiarce?\newline Przyznam się, że gdy kupiłem swoją pierwszą kawiarkę, to kompletnie nie wiedziałem co mam z nią zrobić. Jest to natomiast fantastyczne, a przede wszystkim tanie naczynie, które pozwala przygotować bardzo dobrą kawę.\newline Zasada działania jest prosta. Pod wpływem temperatury woda w dolnym zbiorniku przelewa się pod ciśnieniem przez kawę w pojemniku powyżej i ostatecznie trafia w postaci docelowego napoju do górnego zbiornika. (...)\end{adjustwidth}
)\vspace{5px}\\
\hline
\vspace{5pt}\includegraphics[width=4cm]{questions/sound.png}\vspace{5pt} \newline \emph{\textbf{File description: }A recording of the communication between the Jan Heweliusz ferry and the Mikołaj Kopernik ferry from January 14, 1993.} & \vspace{5px}\textbf{Category:} ASR\vspace{5px}\newline\textbf{Instruction:} [General prompt for ASR]\vspace{5px}\newline\textbf{Verification rules:} \textbf{WAcc}(
\begin{adjustwidth}{0.5cm}{0cm}
- Kopernik, Kopernik - Heweliusz\newline - Tak, odpowiadamy.\newline - Mamy przechył trzydzieści stopni, ogłoszono alarm szalupowy.\newline - Podajcie pozycję.\newline - Podaj pozycję\newline - Heweliusz, Heweliusz - Kopernik\newline - Gdzie jesteście? Czy będziecie, czy pomocy potrzebujecie?\newline - Podajcie pozycję\newline - Jesteście już na, na, na północ, czy na południe od Arkony?\newline - Spróbujcie aktywnym na drugą burtę, żeby wiatr was podniósł.\newline - Spróbujcie aktywnym na drugą burtę, może wiatr was podniesie. (...)\end{adjustwidth}
)\vspace{5px}\\
\hline
\vspace{5pt}\includegraphics[width=4cm]{questions/sound.png}\vspace{5pt} \newline \emph{\textbf{File description: }A recording featuring a monologue in the Silesian dialect discussing ways to spend free time.} & \vspace{5px}\textbf{Category:} Speech QA\vspace{5px}\newline\textbf{Instruction:} Nagranie zawiera monolog w gwarze śląskiej. Na podstawie treści nagrania określ, które z poniższych zdań dotyczących jego treści są prawdziwe:\newline A. Nikaus, autor nagrania, jest doktorem habilitowanym.\newline B. Do ochrony przed słońcem autor poleca parasol albo posmarowanie się kremem.\newline C. Na grilla Niklaus poleca kaszankę albo kiełbasę.\newline D. Autor sugeruje kąpiel w jeziorze na golasa.\newline E. Mówi, że ostatecznie lepiej wakacje spędzić w domu niż nad wodą.\newline F. Ostrzega, aby nie przesadzić z alkoholem.\newline Wypisz tylko listę liter po przecinku odpowiadającym prawdziwym stwierdzeniom, bez dodatkowych komentarzy.\vspace{5px}\newline\textbf{Instruction (english translation):} The recording contains a monologue in the Silesian dialect. Based on the content of the recording, determine which of the following statements regarding its content are true:\newline A. Nikaus, the author of the recording, is a habilitated doctor.\newline B. For sun protection, the author recommends an umbrella or applying sunscreen.\newline C. For a barbecue, Niklaus recommends blood sausage or Polish sausage.\newline D. The author suggests swimming naked in the lake.\newline E. He says that ultimately it is better to spend the holidays at home than by the water.\newline F. He warns not to overdo it with alcohol.\newline Provide only a comma-separated list of letters corresponding to the true statements, without any additional comments.\vspace{5px}\newline\textbf{Verification rules:} \textbf{Include}(B, C, F), \textbf{Exclude}(A, D, E)\vspace{5px}\\
\hline
\vspace{5pt}\includegraphics[width=4cm]{questions/sound.png}\vspace{5pt} \newline \emph{\textbf{File description: }A recording presenting a short excerpt from the biography of the Polish inventor Jan Czochralski.} & \vspace{5px}\textbf{Category:} Speech QA\vspace{5px}\newline\textbf{Instruction:} Podaj imię i nazwisko wynalazcy, o którym mowa w nagraniu. Do otrzymywania czego służy technika, którą odkrył, a która rozsławiła jego nazwisko? \vspace{5px}\newline\textbf{Instruction (english translation):} Provide the first and last name of the inventor mentioned in the recording. What is the technique he discovered, which made his name famous, used to obtain?\vspace{5px}\newline\textbf{Verification rules:} \textbf{Include}([Jan Czochralski, Janie Czochralskim], [monokryształ, monokryształów])\vspace{5px}\\
\hline
\vspace{5pt}\includegraphics[width=4cm]{questions/sound.png}\vspace{5pt} \newline \emph{\textbf{File description: }An excerpt from a song performed live by a Kashubian folk band.} & \vspace{5px}\textbf{Category:} Sound and music QA\vspace{5px}\newline\textbf{Instruction:} Nagranie zawiera fragment polskiej muzyki ludowej. W utworze można usłyszeć ludowy instrument perkusyjny charakterystyczny dla konkretnego regionu Polski. Co to za region i jak nazywa się ten instrument?\vspace{5px}\newline\textbf{Instruction (english translation):} The recording contains an excerpt of Polish folk music. In the piece, you can hear a folk percussion instrument characteristic of a specific region of Poland. What is this region and what is the name of this instrument?\vspace{5px}\newline\textbf{Verification rules:} \textbf{Include}([kaszuby, kaszubska], [skrzypce diabelskie, diabelskie skrzypce])\vspace{5px}\\
\hline
\vspace{5pt}\includegraphics[width=4cm]{questions/sound.png}\vspace{5pt} \newline \emph{\textbf{File description: }Sheep bleating and the sounds of shepherd bells.} & \vspace{5px}\textbf{Category:} Sound and music QA\vspace{5px}\newline\textbf{Instruction:} Wypasowi jakiego rodzaju zwierząt towarzyszą te dźwięki? Jakie regiony Polski są znane z hodowli tych zwierząt?\vspace{5px}\newline\textbf{Instruction (english translation):} The grazing of what kind of animals is accompanied by these sounds? Which regions of Poland are known for breeding these animals?\vspace{5px}\newline\textbf{Verification rules:} \textbf{Include}([owca, owce, owiec], [podhale, podhalański], [beskid, beskidy, beskidzki], [bieszczady, bieszczadzki])\vspace{5px}\\
\hline
\vspace{5px}\includegraphics[width=4cm]{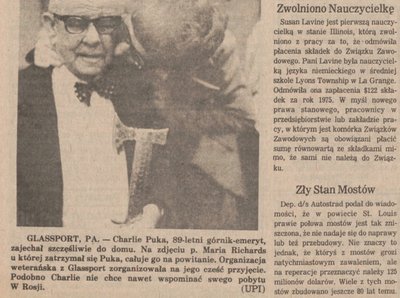} & \vspace{5px}\textbf{Category:} OCR\vspace{5px}\newline\textbf{Instruction:} [General prompt for OCR]\vspace{5px}\newline\textbf{Verification rules:} \textbf{OCR}(\newline\textbf{Block[1]}(
\begin{adjustwidth}{0.50cm}{0cm}
Zwolniono Nauczycielkę\newline Susan Lavine jest pierwszą nauczycielką w stanie Illinois, którą zwolniono z pracy za to, że odmówiła płacenia składek do Związku Zawodowego. Pani Lavine była nauczycielką języka niemieckiego w średniej szkole Lyons Township w La Grange. Odmówiła ona zapłacenia \$122 składek za rok 1975. W myśl nowego prawa stanowego, pracownicy w przedsiębiorstwie lub zakładzie pracy, w którym jest komórka Związków Zawodowych są obowiązani płacić sumę równowartą ze składkami mimo, że sami nie należą do Związku.\end{adjustwidth}
)\newline\textbf{Block[2]}(
\begin{adjustwidth}{0.50cm}{0cm}
Zły Stan Mostów\newline Dep. d/s Autostrad podał do wiadomości, że w powiecie St. Louis prawie połowa mostów jest tak zniszczona, że nie nadaje się do naprawy lub też przebudowy. Nie znaczy to jednak, że któryś z mostów grozi natychmiastowym zawaleniem, ale na reperacje przeznaczyć należy 125 milionów dolarów. Wiele z tych mostów zbudowano jeszcze 80 lat temu.\end{adjustwidth}
)\newline\textbf{Block[3]}(
\begin{adjustwidth}{0.50cm}{0cm}
GLASSPORT, PA. — Charlie Puka, 89-letni górnik-emeryt, zajechał szczęśliwie do domu. Na zdjęciu p. Maria Richards u której zatrzymał się Puka, całuje go na powitanie. Organizacja weterańska z Glassport zorganizowała na jego cześć przyjęcie. Podobno Charlie nie chce nawet wspominać swego pobytu W Rosji.\newline (UPI)\end{adjustwidth}
))\vspace{5px}\\
\hline
\vspace{5px}\includegraphics[width=4cm]{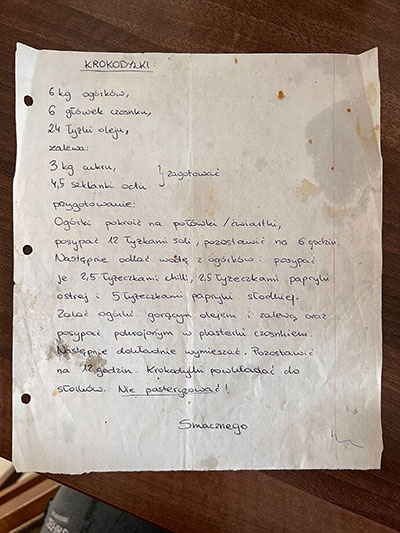} & \vspace{5px}\textbf{Category:} OCR\vspace{5px}\newline\textbf{Instruction:} [General prompt for OCR]\vspace{5px}\newline\textbf{Verification rules:} \textbf{OCR}(
\begin{adjustwidth}{0.50cm}{0cm}
\# KROKODYLKI :\newline 6 kg ogórków,\newline 6 główek czosnku,\newline 24 łyżki oleju,\newline zalewa:\newline 3 kg cukru,\newline 4,5 szklanki octu \newline zagotować\newline przygotowanie:\newline Ogórki pokroić na połówki / ćwiartki, posypać 12 łyżkami soli, pozostawić na 6 godzin. Następnie odlać wodę z ogórków i posypać je 2,5 łyżeczkami chilli, 2,5 łyżeczkami papryki ostrej i 5 łyżeczkami papryki słodkiej. Zalać ogórki gorącym olejem i zalewą oraz posypać pokrojonym w plasterki czosnkiem. Następnie dokładnie wymieszać. Pozostawić na 12 godzin. Krokodylki powkładać do słoików. Nie pasteryzować !\newline Smacznego\end{adjustwidth}
)\vspace{5px}\\
\hline
\vspace{5px}\includegraphics[width=4cm]{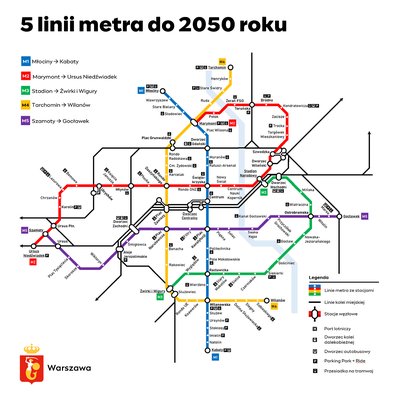} & \vspace{5px}\textbf{Category:} Document QA\vspace{5px}\newline\textbf{Instruction:} Na podstawie załączonego schematu, podaj kolejno nazwy stacji:\newline a) Na której mogę się przesiąść z linii niebieskiej do fioletowej\newline b) Na której mogę wsiąść do trzech różnych linii\newline c) Najbardziej wysuniętej na wschód\newline d) Kiedy jadę czerwoną linią i zarówno od linii żółtej jak i niebieskiej dzieli mnie tylko jeden przystanek\newline Podaj same nazwy stacji w kolejności od a) do d), bez dodatkowych komentarzy.\vspace{5px}\newline\textbf{Instruction (english translation):} Based on the attached map, provide the names of the stations in order:\newline a) Where I can transfer from the blue line to the purple line\newline b) Where I can board three different lines\newline c) The easternmost one\newline d) When I am traveling on the red line and am only one stop away from both the yellow and blue lines\newline Provide only the names of the stations in order from a) to d), without any additional comments.\vspace{5px}\newline\textbf{Verification rules:} \textbf{Order}(Plac Konstytucji, Marymont, Gocławek, Rondo ONZ)\vspace{5px}\\
\hline
\vspace{5px}\includegraphics[width=4cm]{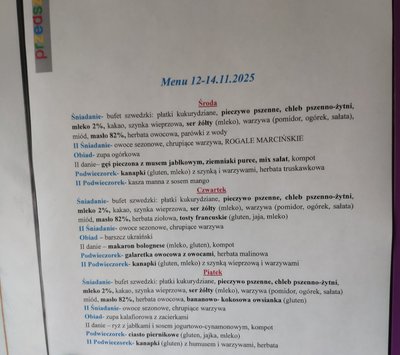} & \vspace{5px}\textbf{Category:} Document QA\vspace{5px}\newline\textbf{Instruction:} Na podstawie informacji zawartych w załączonym zdjęciu menu z przedszkola odpowiedz na poniższe pytania.\newline 1. Jakie smaki herbat dostają dzieci?\newline 2. Z jakim świętem państwowym związany jest przysmak, który dzieci będą jadły w środę na drugie śniadanie?\newline 3. Jakie produkty zawierają gluten patrząc jedynie na informacje w nawiasach? \newline Odpowiedz tylko na pytania, bez dodatkowego komentarza.\vspace{5px}\newline\textbf{Instruction (english translation):} Based on the information contained in the attached photo of the kindergarten menu, answer the following questions.\newline 1. What flavors of tea do the children get?\newline 2. What national holiday is associated with the treat the children will eat for second breakfast on Wednesday?\newline 3. Which products contain gluten looking only at the information in parentheses? \newline Answer only the questions, without any additional comments.\vspace{5px}\newline\textbf{Verification rules:} \textbf{Include}(owocowa, truskawkowa, ziołowa, malinowa, [Dzień Niepodległości, Święto Niepodległości], kanapki, tosty francuskie, makaron bolognese, [bananowo-kokosowa owsianka, bananowo kokosowa owsianka], ciasto piernikowe), \textbf{Exclude}(pieczywo pszenne, chleb pszenno-żytni)\vspace{5px}\\
\hline
\vspace{5px}\includegraphics[width=4cm]{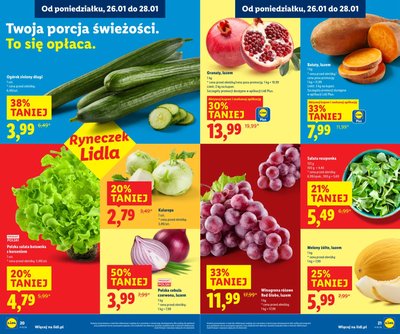} & \vspace{5px}\textbf{Category:} Structured extraction\vspace{5px}\newline\textbf{Instruction:} [General prompt for Structured extraction] + Na podstawie fragmentu gazetki promocyjnej dyskontu spożywczego, wyodrębnij informacje o wszystkich produktach, które są przecenione o 30\% lub więcej. Zwróć nazwę produktu, jego cenę przed obniżką i po obniżce. Odpowiedz w postaci listy obiektów JSON.\vspace{5px}\newline\textbf{Instruction (english translation):} [General prompt for Structured extraction] + Based on an excerpt from a discount grocery store's promotional flyer, extract information about all products that are discounted by 30\% or more. Return the product name, its price before and after the discount. Provide the answer as a list of JSON objects.\vspace{5px}\newline\textbf{JSON schema:}
\{\textcolor[HTML]{A71D5D}{"name"}: \textcolor[HTML]{183691}{"Object"}, \textcolor[HTML]{A71D5D}{"schema"}: \{\textcolor[HTML]{A71D5D}{"type"}: \textcolor[HTML]{183691}{"array"}, \textcolor[HTML]{A71D5D}{"items"}: \{\textcolor[HTML]{A71D5D}{"type"}: \textcolor[HTML]{183691}{"object"}, \textcolor[HTML]{A71D5D}{"properties"}: \{\textcolor[HTML]{A71D5D}{"product"}: \{\textcolor[HTML]{A71D5D}{"type"}: \textcolor[HTML]{183691}{"string"}, \textcolor[HTML]{A71D5D}{"description"}: \textcolor[HTML]{183691}{"Nazwa produktu"}\}, \textcolor[HTML]{A71D5D}{"before"}: \{\textcolor[HTML]{A71D5D}{"type"}: \textcolor[HTML]{183691}{"number"}, \textcolor[HTML]{A71D5D}{"description"}: \textcolor[HTML]{183691}{"Cena przed obniżką"}\}, \textcolor[HTML]{A71D5D}{"after"}: \{\textcolor[HTML]{A71D5D}{"type"}: \textcolor[HTML]{183691}{"number"}, \textcolor[HTML]{A71D5D}{"description"}: \textcolor[HTML]{183691}{"Cena po obniżce"}\}\}, \textcolor[HTML]{A71D5D}{"required"}: [\textcolor[HTML]{183691}{"product"}, \textcolor[HTML]{183691}{"before"}, \textcolor[HTML]{183691}{"after"}], \textcolor[HTML]{A71D5D}{"additionalProperties"}: \textcolor[HTML]{D73A49}{false}\}\}\}\vspace{5px}\newline
\textbf{Verification rules: Structure(}
\begin{adjustwidth}{0.50cm}{0cm}
[\{\textcolor[HTML]{A71D5D}{"product"}: \textcolor[HTML]{183691}{"Ogórek zielony długi"}, \textcolor[HTML]{A71D5D}{"before"}: \textcolor[HTML]{0086B3}{6.49}, \textcolor[HTML]{A71D5D}{"after"}: \textcolor[HTML]{0086B3}{3.99}\}, \{\textcolor[HTML]{A71D5D}{"product"}: \textcolor[HTML]{183691}{"Granaty, luzem"}, \textcolor[HTML]{A71D5D}{"before"}: \textcolor[HTML]{0086B3}{19.99}, \textcolor[HTML]{A71D5D}{"after"}: \textcolor[HTML]{0086B3}{13.99}\}, \{\textcolor[HTML]{A71D5D}{"product"}: \textcolor[HTML]{183691}{"Bataty, luzem"}, \textcolor[HTML]{A71D5D}{"before"}: \textcolor[HTML]{0086B3}{11.99}, \textcolor[HTML]{A71D5D}{"after"}: \textcolor[HTML]{0086B3}{7.99}\}, \{\textcolor[HTML]{A71D5D}{"product"}: \textcolor[HTML]{183691}{"Polska cebula czerwona, luzem"}, \textcolor[HTML]{A71D5D}{"before"}: \textcolor[HTML]{0086B3}{7.99}, \textcolor[HTML]{A71D5D}{"after"}: \textcolor[HTML]{0086B3}{3.99}\}, \{\textcolor[HTML]{A71D5D}{"product"}: \textcolor[HTML]{183691}{"Winogrona różowe Red Globe, luzem"}, \textcolor[HTML]{A71D5D}{"before"}: \textcolor[HTML]{0086B3}{17.99}, \textcolor[HTML]{A71D5D}{"after"}: \textcolor[HTML]{0086B3}{11.99}\}]\end{adjustwidth}
)\vspace{5px}\\
\hline
\vspace{5px}\includegraphics[width=4cm]{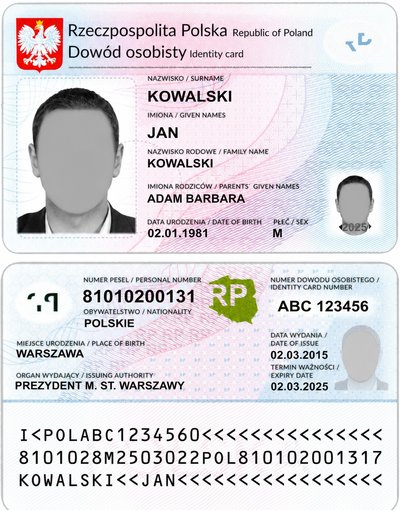} & \vspace{5px}\textbf{Category:} Structured extraction\vspace{5px}\newline\textbf{Instruction:} [General prompt for Structured extraction] + Wyodrębnij informacje zawarte na przykładowym dokumencie tożsamości. Odpowiedź zwróć w formacie JSON.\vspace{5px}\newline\textbf{Instruction (english translation):} [General prompt for Structured extraction] + Extract the information contained on the sample identity document. Return the response in JSON format.\vspace{5px}\newline\textbf{JSON schema:}
\{\textcolor[HTML]{A71D5D}{"name"}: \textcolor[HTML]{183691}{"Object"}, \textcolor[HTML]{A71D5D}{"schema"}: \{\textcolor[HTML]{A71D5D}{"type"}: \textcolor[HTML]{183691}{"object"}, \textcolor[HTML]{A71D5D}{"properties"}: \{\textcolor[HTML]{A71D5D}{"NAZWISKO"}: \{\textcolor[HTML]{A71D5D}{"type"}: \textcolor[HTML]{183691}{"string"}, \textcolor[HTML]{A71D5D}{"description"}: \textcolor[HTML]{183691}{"Nazwisko"}\}, \textcolor[HTML]{A71D5D}{"IMIONA"}: \{\textcolor[HTML]{A71D5D}{"type"}: \textcolor[HTML]{183691}{"string"}, \textcolor[HTML]{A71D5D}{"description"}: \textcolor[HTML]{183691}{"Imiona"}\}, \textcolor[HTML]{A71D5D}{"NAZWISKO RODOWE"}: \{\textcolor[HTML]{A71D5D}{"type"}: \textcolor[HTML]{183691}{"string"}, \textcolor[HTML]{A71D5D}{"description"}: \textcolor[HTML]{183691}{"Nazwisko rodowe"}\}, \textcolor[HTML]{A71D5D}{"IMIONA RODZICÓW"}: \{\textcolor[HTML]{A71D5D}{"type"}: \textcolor[HTML]{183691}{"string"}, \textcolor[HTML]{A71D5D}{"description"}: \textcolor[HTML]{183691}{"Imiona rodziców"}\}, \textcolor[HTML]{A71D5D}{"DATA URODZENIA"}: \{\textcolor[HTML]{A71D5D}{"type"}: \textcolor[HTML]{183691}{"string"}, \textcolor[HTML]{A71D5D}{"description"}: \textcolor[HTML]{183691}{"Data urodzenia"}\}, \textcolor[HTML]{A71D5D}{"PŁEĆ"}: \{\textcolor[HTML]{A71D5D}{"type"}: \textcolor[HTML]{183691}{"string"}, \textcolor[HTML]{A71D5D}{"description"}: \textcolor[HTML]{183691}{"Płeć"}\}, \textcolor[HTML]{A71D5D}{"NUMER PESEL"}: \{\textcolor[HTML]{A71D5D}{"type"}: \textcolor[HTML]{183691}{"string"}, \textcolor[HTML]{A71D5D}{"description"}: \textcolor[HTML]{183691}{"Numer PESEL"}\}, \textcolor[HTML]{A71D5D}{"OBYWATELSTWO"}: \{\textcolor[HTML]{A71D5D}{"type"}: \textcolor[HTML]{183691}{"string"}, \textcolor[HTML]{A71D5D}{"description"}: \textcolor[HTML]{183691}{"Obywatelstwo"}\}, \textcolor[HTML]{A71D5D}{"MIEJSCE URODZENIA"}: \{\textcolor[HTML]{A71D5D}{"type"}: \textcolor[HTML]{183691}{"string"}, \textcolor[HTML]{A71D5D}{"description"}: \textcolor[HTML]{183691}{"Miejsce urodzenia"}\}, \textcolor[HTML]{A71D5D}{"ORGAN WYDAJĄCY"}: \{\textcolor[HTML]{A71D5D}{"type"}: \textcolor[HTML]{183691}{"string"}, \textcolor[HTML]{A71D5D}{"description"}: \textcolor[HTML]{183691}{"Organ wydający"}\}, \textcolor[HTML]{A71D5D}{"NUMER DOWODU OSOBISTEGO"}: \{\textcolor[HTML]{A71D5D}{"type"}: \textcolor[HTML]{183691}{"string"}, \textcolor[HTML]{A71D5D}{"description"}: \textcolor[HTML]{183691}{"Numer dowodu osobistego"}\}, \textcolor[HTML]{A71D5D}{"DATA WYDANIA"}: \{\textcolor[HTML]{A71D5D}{"type"}: \textcolor[HTML]{183691}{"string"}, \textcolor[HTML]{A71D5D}{"description"}: \textcolor[HTML]{183691}{"Data wydania"}\}, \textcolor[HTML]{A71D5D}{"TERMIN WAŻNOŚCI"}: \{\textcolor[HTML]{A71D5D}{"type"}: \textcolor[HTML]{183691}{"string"}, \textcolor[HTML]{A71D5D}{"description"}: \textcolor[HTML]{183691}{"Termin ważności"}\}\}, \textcolor[HTML]{A71D5D}{"required"}: [\textcolor[HTML]{183691}{"NAZWISKO"}, \textcolor[HTML]{183691}{"IMIONA"}, \textcolor[HTML]{183691}{"NAZWISKO RODOWE"}, \textcolor[HTML]{183691}{"IMIONA RODZICÓW"}, \textcolor[HTML]{183691}{"DATA URODZENIA"}, \textcolor[HTML]{183691}{"PŁEĆ"}, \textcolor[HTML]{183691}{"NUMER PESEL"}, \textcolor[HTML]{183691}{"OBYWATELSTWO"}, \textcolor[HTML]{183691}{"MIEJSCE URODZENIA"}, \textcolor[HTML]{183691}{"ORGAN WYDAJĄCY"}, \textcolor[HTML]{183691}{"NUMER DOWODU OSOBISTEGO"}, \textcolor[HTML]{183691}{"DATA WYDANIA"}, \textcolor[HTML]{183691}{"TERMIN WAŻNOŚCI"}], \textcolor[HTML]{A71D5D}{"additionalProperties"}: \textcolor[HTML]{D73A49}{false}\}\}\vspace{5px}\newline
\textbf{Verification rules: Structure(}
\begin{adjustwidth}{0.50cm}{0cm}
\{\textcolor[HTML]{A71D5D}{"NAZWISKO"}: \textcolor[HTML]{183691}{"KOWALSKI"}, \textcolor[HTML]{A71D5D}{"IMIONA"}: \textcolor[HTML]{183691}{"JAN"}, \textcolor[HTML]{A71D5D}{"NAZWISKO RODOWE"}: \textcolor[HTML]{183691}{"KOWALSKI"}, \textcolor[HTML]{A71D5D}{"IMIONA RODZICÓW"}: \textcolor[HTML]{183691}{"ADAM BARBARA"}, \textcolor[HTML]{A71D5D}{"DATA URODZENIA"}: \textcolor[HTML]{183691}{"02.01.1981"}, \textcolor[HTML]{A71D5D}{"PŁEĆ"}: \textcolor[HTML]{183691}{"M"}, \textcolor[HTML]{A71D5D}{"NUMER PESEL"}: \textcolor[HTML]{183691}{"81010200131"}, \textcolor[HTML]{A71D5D}{"OBYWATELSTWO"}: \textcolor[HTML]{183691}{"POLSKIE"}, \textcolor[HTML]{A71D5D}{"MIEJSCE URODZENIA"}: \textcolor[HTML]{183691}{"WARSZAWA"}, \textcolor[HTML]{A71D5D}{"ORGAN WYDAJĄCY"}: \textcolor[HTML]{183691}{"PREZYDENT M. ST. WARSZAWY"}, \textcolor[HTML]{A71D5D}{"NUMER DOWODU OSOBISTEGO"}: \textcolor[HTML]{183691}{"ABC 123456"}, \textcolor[HTML]{A71D5D}{"DATA WYDANIA"}: \textcolor[HTML]{183691}{"02.03.2015"}, \textcolor[HTML]{A71D5D}{"TERMIN WAŻNOŚCI"}: \textcolor[HTML]{183691}{"02.03.2025"}\}\end{adjustwidth}
)\vspace{5px}\\
\hline
\end{longtable}
}

\end{document}